\documentclass[1p,12pt]{elsarticle}
\biboptions{sort&compress}

\usepackage{amssymb}
\usepackage{placeins}
\usepackage{amsmath}
\usepackage{graphicx}%
\usepackage{multirow}%
\usepackage{amsmath,amssymb,amsfonts}%
\usepackage{amsthm}%
\usepackage{mathrsfs}%
\usepackage[title]{appendix}%
\usepackage{xcolor}%
\usepackage{textcomp}%
\usepackage{manyfoot}%
\usepackage{booktabs}%
\usepackage{algorithm}%
\usepackage{algorithmicx}%
\usepackage{algpseudocode}%
\usepackage{listings}%

\usepackage{tikz}
\usetikzlibrary{positioning,arrows.meta,backgrounds,calc,decorations.pathreplacing}

\usepackage{hyperref}
\usepackage{geometry}
\usepackage{tabularx} 
\usepackage{microtype}
\usepackage{subcaption}

\theoremstyle{thmstyleone}%

\theoremstyle{thmstyletwo}%

\theoremstyle{thmstylethree}%

\begin{document}

\begin{frontmatter}

\title{Spiking Neural Networks for CT-Based Liver Lesion Presence Classification Under Leakage-Free Patient-Level Evaluation: A Two-Dataset Benchmark} 

\author[1]{Zofia Rudnicka}
\author[1]{Janusz Szczepanski}
\author[1]{Agnieszka Pregowska\corref{cor1}}

\cortext[cor1]{Corresponding author}
\ead{aprego@ippt.pan.pl}

\affiliation[1]{
  organization={Institute of Fundamental Technological Research, Polish Academy of Sciences},
  addressline={Pawinskiego 5B},
  city={Warsaw},
  postcode={02-106},
  country={Poland}
}

\begin{abstract}
\textbf{Purpose:} Spiking neural networks (SNNs) offer an event-driven and biologically inspired alternative to conventional deep learning, yet their potential in abdominal CT analysis remains largely unexplored. This study aims to provide a systematic benchmark of SNN architectures for liver lesion presence classification in CT imaging and to assess their performance relative to a conventional CNN baseline under a strictly leakage-free patient-level evaluation protocol.

\textbf{Methods:} We evaluated leakage-free patient-level liver lesion presence classification on two abdominal CT benchmarks: the \texttt{Task03\_Liver} dataset from the Medical Segmentation Decathlon, comprising 131 contrast-enhanced abdominal CT volumes with expert liver and lesion annotations, and an additional contrast-enhanced CT dataset (\texttt{Task03\_CECT}) processed under the same patient-level evaluation protocol. Slice- and patient-level labels were derived directly from the lesion masks, while all train/validation/test splits were performed strictly at the patient level. We compared a conventional CNN baseline with three spiking approaches: a convolutional surrogate-gradient SNN (ConvSNN), an attention-based ConvSNN multiple-instance learning model (ConvSNN-MIL), and a Tempotron-style classifier with time-to-first-spike encoding. Due to class imbalance, performance was assessed primarily using patient-level precision-recall AUC (PR-AUC), together with ROC-AUC, F1-score, accuracy, and Matthews correlation coefficient.


\textbf{Results:} Across five independent leakage-free patient-level splits, the CNN baseline achieved the highest mean PR-AUC on the \texttt{Task03\_Liver} benchmark ($0.970 \pm 0.036$), whereas the convolutional surrogate-gradient SNN achieved the strongest threshold-dependent patient-level performance, including the highest mean accuracy, F1-score, and MCC. On the additional \texttt{Task03\_CECT} benchmark, performance decreased for all models, but the convolutional spike-based models remained competitive, with ConvSNN-MIL outperforming the standard ConvSNN on several patient-level metrics.

\textbf{Conclusion:} These results suggest that carefully designed convolutional spiking neural networks, evaluated under strict leakage-free patient-level protocols, can achieve performance comparable to a strong CNN baseline and may retain competitive performance across distinct abdominal CT cohorts processed under the same evaluation protocol. The study further underlines the importance of patient-level evaluation, convolutional spike-based feature extraction, and learning-rule selection when assessing the methodological utility of SNNs in medical imaging.

\end{abstract}

\begin{keyword}
Spiking Neural Networks (SNN) \sep Liver Lesion Presence Classification \sep Computed Tomography (CT) \sep Patient-Level Evaluation \sep Temporal Encoding \sep Medical Image Analysis


\end{keyword}

\end{frontmatter}



\section{Introduction}\label{sec1}

Liver diseases remain among the leading causes of morbidity and mortality worldwide, with hepatocellular carcinoma, cirrhosis, and non-alcoholic fatty liver disease accounting for a substantial and growing global healthcare burden \cite{guo2025}. Early and reliable detection of hepatic pathology is therefore essential to improve therapeutic outcomes, enable timely clinical intervention, and limit disease progression \cite{pozowski2025}. In this context, medical imaging has become a cornerstone of noninvasive liver assessment, with computed tomography (CT) playing a central role due to its wide availability, high spatial resolution, and robustness in routine clinical workflows \cite{hu2023}. 

Despite these advantages, manual interpretation of abdominal CT scans is time-consuming, prone to variability between observers, and increasingly constrained by the limited availability of experienced radiologists \cite{chupetlovska2025}. These challenges have driven rapid progress in machine learning-based diagnostic systems that support radiological decision making. However, a critical methodological limitation persists in much of the existing literature: many deep learning approaches report performance at the slice or patch level, implicitly assuming independence between samples extracted from the same patient volume. In volumetric CT data, such assumptions are violated due to strong intra-patient correlations, leading to information leakage and systematic overestimation of clinical performance. This motivates the adoption of strictly patient-level evaluation protocols when assessing diagnostic models for liver CT.

Convolutional neural networks (CNNs) currently dominate liver lesion detection and classification tasks, achieving strong performance across a range of imaging modalities. Nevertheless, their dense, synchronous computation and limited capacity for explicit temporal modeling pose challenges in terms of computational efficiency and biological interpretability, particularly for volumetric and sequential data \cite{ying2024,he2024}. These limitations highlight the potential of bio-inspired approaches. Spiking Neural Networks (SNNs), which operate using event-driven computation and temporally coded information, offer a promising yet underexplored alternative for CT-based liver analysis \cite{xue2023,yang2023}. By representing information through temporally coded spike trains and neuronal dynamics, SNNs provide an alternative computational paradigm for processing high-dimensional image representations.

From a theoretical perspective, spiking neural networks introduce an additional computational dimension compared with conventional feedforward neural networks, namely the temporal dynamics of neuronal states. In LIF-based architectures, membrane potentials integrate inputs over time and generate spikes only when sufficient evidence accumulates. This temporal integration can be interpreted as a form of dynamic evidence accumulation, which may improve robustness to noise and local variability in high-dimensional medical images. In volumetric CT data, where anatomical structures extend across multiple adjacent slices and subtle lesions may occupy only small regions, such temporally integrated representations may help stabilize feature extraction and emphasize spatially consistent patterns.

Recent studies have demonstrated the applicability of SNNs to medical image classification, though predominantly in non-abdominal domains such as pulmonary, neurological, dermatological, breast, and thermographic imaging \cite{doborjeh2022,rudnicka2026}. For instance, Tempotron-based SNNs augmented with spike-timing-dependent plasticity have achieved accuracies exceeding 98.00\% in tuberculosis detection from chest X-ray datasets \cite{patankar2025}. Similarly, surrogate gradient-trained SNNs have been applied to melanoma classification in dermoscopy images with reported accuracies near 89.60\% \cite{gilani2023}, while multispike architectures with temporal feedback have achieved up to 98.00\% accuracy in breast cancer diagnosis \cite{heidarian2024}. 

Beyond organ-specific applications, several efforts have aimed to construct more generalizable SNN frameworks. The SNMID system, for example, combined wavelet and histogram-based features with spiking models to classify diverse medical conditions, including malaria, breast cancer, and skin lesions, achieving competitive performance \cite{dehariya2021}. However, such approaches rely heavily on handcrafted features and do not fully exploit temporal spike-based learning, limiting scalability and adaptability. Importantly, despite this growing body of work, the application of SNNs to abdominal CT, particularly liver lesion analysis, remains largely unexplored, representing a clear methodological and translational gap \cite{indiveri2009}.

This study addresses this gap by presenting a comprehensive benchmark of spiking neural networks for binary liver lesion presence classification from abdominal CT imaging. To the best of our knowledge, relatively few studies have systematically evaluated spiking neural networks for abdominal CT liver lesion presence classification under a strictly patient-level protocol. In our formulation, a patient case is considered lesion-present if at least one axial slice contains lesion pixels in the expert annotation (original label~2), yielding a clinically meaningful binary endpoint: lesion-absent versus lesion-present. 

To systematically investigate spike-based learning in this setting, we introduce the SNNDeep framework, a lightweight spiking neural network implementation designed specifically for hepatic CT analysis. Unlike many existing SNN approaches that rely on high-level abstractions provided by open-source frameworks, SNNDeep is implemented from the ground up, enabling fine-grained control over neuron dynamics, spike generation, temporal encoding, and synaptic updates. This low-level design facilitates systematic investigation of biologically inspired learning rules and architectural variants \cite{rudnicka2026neuron}.

To rigorously assess the impact of architectural design and learning strategy, we benchmark a conventional CNN baseline against three spiking approaches operating under the same leakage-free patient-level evaluation protocol: a convolutional SNN trained with surrogate gradients (ConvSNN), an attention-based ConvSNN multiple-instance learning variant (ConvSNN-MIL), and a Tempotron-style model trained with time-to-first-spike encoding. All models are evaluated under strict patient-level splitting to prevent information leakage, using the Task03\_Liver dataset from the Medical Segmentation Decathlon as the primary benchmark and an additional contrast-enhanced CT dataset (Task03\_CECT) for cross-dataset validation. Because of substantial class imbalance, performance is reported primarily using patient-level PR-AUC, together with ROC-AUC, F1-score, accuracy, and Matthews correlation coefficient. In addition to methodological benchmarking, this work aims to provide insights into the practical applicability of spiking neural networks in clinically relevant CT-based diagnostic scenarios.

\section{Materials and Methods}

\subsection{Datasets and preprocessing}

Two abdominal CT datasets were used in this study to evaluate the robustness of spike-based models under leakage-free patient-level classification protocols.

\paragraph{Dataset 1: \texttt{Task03\_Liver}}
The first dataset was the \texttt{Task03\_Liver} dataset from the Medical Segmentation Decathlon (MSD) \cite{antonelli2022}, which contains 131 contrast-enhanced abdominal CT volumes with expert annotations for liver (label~1) and liver lesions (label~2). No segmentation model was trained; instead, the provided masks were used exclusively to derive binary lesion-presence labels.

For each patient, the total number of lesion voxels was computed from the lesion mask (label~2). A patient was labeled as lesion-positive if at least one voxel belonging to a liver lesion was present in the expert annotation. Thus, the binary endpoint corresponds to the presence or absence of liver lesions at the patient level.

\paragraph{Dataset 2: \texttt{Task03\_CECT}}
The second dataset, denoted \texttt{Task03\_CECT}, was constructed as an additional contrast-enhanced abdominal CT benchmark using the same binary patient-level lesion-presence formulation \cite{luo2025}. This dataset differs from \texttt{Task03\_Liver} in acquisition source, cohort composition, and imaging distribution, and was included to test whether the observed model ranking and patient-level performance trends remained stable under a distinct abdominal CT setting. For this dataset, patient-level binary labels were derived using the same lesion-presence criterion as for \texttt{Task03\_Liver}, i.e., a case was labeled positive if at least one annotated liver lesion was present within the volume. All preprocessing, patient-level splitting, and evaluation procedures were kept identical to those used for the MSD benchmark to ensure methodological comparability across datasets.

For consistency, the binary patient label in both datasets indicated whether at least one liver lesion was present within the annotated volume.

The voxel-level annotations were used exclusively to derive patient-level labels and were not provided as inputs to the classification models. This definition reflects a clinically relevant screening scenario in which the objective is to identify patients with any detectable liver lesion.

Rather than using isolated 2D slices, the models were trained on patient-level bags composed of axial slices. Each slice was represented using a 2.5D context with axial offsets $(-2,-1,0,1,2)$ around the reference slice. For each contextual slice, three Hounsfield-unit windows were applied: $[-100,200]$, $[0,300]$, and $[-150,250]$. This yielded a multi-channel representation with $5 \times 3 = 15$ channels per slice. Liver masks (label~1) were used to define a region of interest, and each slice was cropped to the liver bounding box with a margin before being resized to $96 \times 96$ pixels.

The liver masks were used solely to define a region of interest (ROI) for cropping in order to reduce irrelevant background context. These masks were not used during model inference as predictive inputs but only to restrict the spatial field analyzed by the classifier. In practical clinical deployment, the same preprocessing step could be replaced by an automated liver segmentation model. Accordingly, the present experiments should be interpreted as evaluating lesion-presence classification under a controlled liver-localized setting rather than as a fully end-to-end annotation-free diagnostic pipeline.

For positive patients, training bags were enriched with lesion-containing slices, while negative patients were sampled uniformly from the available axial slices. Lesion annotations were used only during patient-level label derivation and training-time positive-slice enrichment; they were not used during validation or test-time inference. Mild data augmentation, including flips, noise, and gamma perturbations, was applied during training. This preprocessing produced patient-level bags suitable for both convolutional SNN and MIL-based evaluation. In the MIL setting, patient bags were resampled during training to improve slice diversity within each epoch. Unless insufficient slices were available, bag slices were sampled without replacement. During inference, multiple independently sampled bags per patient were evaluated and their predicted probabilities were averaged. The same preprocessing protocol was applied to both datasets unless otherwise stated, ensuring that any performance differences reflected dataset characteristics rather than changes in the processing pipeline.

\subsection{Patient-level splitting and evaluation protocol}

Axial slices from the same CT volume are strongly correlated. To prevent information leakage, the dataset was partitioned strictly at the patient level using repeated stratified random splits over binary patient labels. For each random seed, approximately 60\% of patients were assigned to the training set, 20\% to the validation set, and 20\% to the test set while preserving the patient-level class distribution.

This procedure was repeated for five independent random seeds $\{0,1,2,3,4\}$, resulting in five independent train/validation/test partitions. Performance metrics reported in this work correspond to the mean and standard deviation across these repeated splits.

All model selection, threshold tuning, and aggregation choices were based exclusively on validation patients and then transferred without modification to the corresponding unseen test patients. This protocol ensures that the reported results reflect clinically relevant case-level generalization rather than slice-level memorization.

\subsection{Temporal spike encoding}

For the convolutional SNN models, each multi-channel slice representation was interpreted as an input intensity tensor in $[0,1]$. During training, surrogate-gradient SNNs used stochastic Poisson-like spike generation,
\begin{equation}
S_{t,j} \sim \mathrm{Bernoulli}(x_j),
\end{equation}
where $x_j$ denotes the normalized input activation. During inference, deterministic rate encoding was used,
\begin{equation}
S_{t,j} = x_j, \qquad t=1,\dots,T,
\end{equation}
to reduce stochastic variance. In addition, a Poisson-to-rate annealing strategy was used during training, with early epochs employing stochastic spike generation and later epochs using rate-based inputs for stabilization.

For the Tempotron model, time-to-first-spike (TTFS) encoding was applied. Each scalar input feature generated at most one spike at a latency inversely related to its normalized intensity,
\begin{equation}
t_j = \mathrm{round}\big((1-x_j)(T-1)\big),
\end{equation}
for all $x_j > 0$. The number of simulation steps was set according to the model type, with the convolutional SNN using a longer temporal horizon than the Tempotron model.

\subsection{Neuron model}

All spiking models are based on a discrete-time leaky integrate-and-fire (LIF) neuron model \cite{wu2025}. For neuron $k$ at time step $t$, the membrane potential evolves as
\begin{equation}
V_k[t] = \beta V_k[t-1] + I_k[t],
\end{equation}
where $\beta\in(0,1)$ is the membrane decay factor and $I_k[t]$ denotes the synaptic input. The decay factor $\beta$ models the leakage of the membrane potential over time, reflecting the limited temporal memory of biological neurons. This mechanism introduces a form of temporal filtering, allowing the neuron to accumulate evidence over several time steps while gradually forgetting outdated inputs. A spike is generated whenever the membrane potential exceeds a fixed threshold $V_{\mathrm{th}}$:
\begin{equation}
S_k[t] = \mathbb{I}(V_k[t] - V_{\mathrm{th}} > 0),
\end{equation}
after which the membrane potential is reset according to
\begin{equation}
V_k[t] \leftarrow V_k[t](1 - S_k[t]).
\end{equation}
In all surrogate-gradient SNN experiments, the decay and threshold were fixed to $\beta=0.95$ and $V_{\mathrm{th}}=1.0$ to ensure a controlled comparison across learning rules. 

\begin{figure}[t]
\centering
\resizebox{0.95\textwidth}{!}{
\begin{tikzpicture}[
    font=\sffamily,
    block/.style={
        rounded corners=6pt,
        draw=black!20,
        fill=black!3,
        thick,
        minimum height=11mm,
        minimum width=24mm,
        align=center
    },
    conv/.style={
        rounded corners=6pt,
        draw=green!45!black,
        fill=green!12,
        thick,
        minimum height=11mm,
        minimum width=26mm,
        align=center
    },
    lif/.style={
        rounded corners=6pt,
        draw=violet!55!black,
        fill=violet!12,
        thick,
        minimum height=11mm,
        minimum width=24mm,
        align=center
    },
    head/.style={
        rounded corners=6pt,
        draw=cyan!55!black,
        fill=cyan!14,
        thick,
        minimum height=11mm,
        minimum width=26mm,
        align=center
    },
    arrow/.style={
        -{Stealth[length=2.5mm,width=1.8mm]},
        thick,
        draw=black!60
    },
    note/.style={font=\footnotesize\sffamily, align=center, text=black!70}
]

\node[block] (input) at (0,0)
{Input bag slice\\\textbf{$96\times96\times15$}\\\footnotesize 2.5D + HU windows};

\node[conv] (c1) at (3.5,0)
{ConvLIF 1\\\footnotesize $15\rightarrow24$\\$3\times3$, stride 1};

\node[conv] (c2) at (7.2,0)
{ConvLIF 2\\\footnotesize $24\rightarrow48$\\$3\times3$, stride 2};

\node[conv] (c3) at (10.9,0)
{ConvLIF 3\\\footnotesize $48\rightarrow96$\\$3\times3$, stride 2};

\node[head] (gap) at (14.5,0)
{Global average pooling\\+ linear head};

\node[head] (out) at (18.2,0)
{Patient/slice logit\\\footnotesize $\hat{y}$};

\draw[arrow] (input) -- (c1);
\draw[arrow] (c1) -- (c2);
\draw[arrow] (c2) -- (c3);
\draw[arrow] (c3) -- (gap);
\draw[arrow] (gap) -- (out);

\node[note] at (3.5,-1.5) {LIF neurons\\surrogate gradients};
\node[note] at (7.2,-1.5) {Poisson training\\rate evaluation};
\node[note] at (10.9,-1.5) {Temporal accumulation\\over $T$ steps};
\node[note] at (14.5,-1.5) {Dense readout from\\mean spike activity};

\end{tikzpicture}
}
\caption{Architecture of the convolutional spiking neural network. Each preprocessed CT slice is represented as a $96\times96\times15$ tensor obtained from 2.5D axial context and three HU windows. The input is processed by three convolutional leaky integrate-and-fire blocks and mapped to a single logit through global average pooling and a linear classifier.}
\label{fig:convsnn_arch}
\end{figure}

\begin{figure}[t]
\centering
\resizebox{0.95\textwidth}{!}{
\begin{tikzpicture}[
    font=\sffamily,
    block/.style={
        rounded corners=6pt,
        draw=black!20,
        fill=black!3,
        thick,
        minimum height=11mm,
        minimum width=24mm,
        align=center
    },
    conv/.style={
        rounded corners=6pt,
        draw=green!45!black,
        fill=green!12,
        thick,
        minimum height=11mm,
        minimum width=24mm,
        align=center
    },
    pool/.style={
        rounded corners=6pt,
        draw=orange!60!black,
        fill=orange!14,
        thick,
        minimum height=11mm,
        minimum width=26mm,
        align=center
    },
    head/.style={
        rounded corners=6pt,
        draw=cyan!55!black,
        fill=cyan!14,
        thick,
        minimum height=11mm,
        minimum width=24mm,
        align=center
    },
    arrow/.style={
        -{Stealth[length=2.5mm,width=1.8mm]},
        thick,
        draw=black!60
    },
    brace/.style={
        decorate,
        decoration={brace, amplitude=5pt},
        thick
    },
    note/.style={font=\footnotesize\sffamily, align=center, text=black!70}
]

\node[block] (s1) at (0,1.8) {Slice 1\\$96\times96\times15$};
\node[block] (s2) at (0,0.6) {Slice 2\\$96\times96\times15$};
\node[block] (s3) at (0,-0.6) {$\vdots$};
\node[block] (s4) at (0,-1.8) {Slice $S$\\$96\times96\times15$};

\node[conv] (enc) at (4.2,0)
{Shared ConvSNN encoder\\\footnotesize ConvLIF blocks\\$15\rightarrow24\rightarrow48\rightarrow96$};

\node[head] (f1) at (8.2,1.8) {feat$_1$};
\node[head] (f2) at (8.2,0.6) {feat$_2$};
\node[head] (f3) at (8.2,-0.6) {$\vdots$};
\node[head] (f4) at (8.2,-1.8) {feat$_S$};

\node[pool] (attn) at (12.2,0)
{Attention MIL pooling\\\footnotesize weighted aggregation};

\node[head] (cls) at (15.8,0)
{Linear classifier};

\node[head] (out) at (18.8,0)
{Patient-level\\probability};

\draw[arrow] (s1) -- (enc);
\draw[arrow] (s2) -- (enc);
\draw[arrow] (s3) -- (enc);
\draw[arrow] (s4) -- (enc);

\draw[arrow] (enc) -- (f1);
\draw[arrow] (enc) -- (f2);
\draw[arrow] (enc) -- (f3);
\draw[arrow] (enc) -- (f4);

\draw[arrow] (f1) -- (attn);
\draw[arrow] (f2) -- (attn);
\draw[arrow] (f3) -- (attn);
\draw[arrow] (f4) -- (attn);

\draw[arrow] (attn) -- (cls);
\draw[arrow] (cls) -- (out);

\draw[brace] (-1.2,2.35) -- (-1.2,-2.35)
node[midway,left=6pt,align=center] {\footnotesize Patient bag\\\footnotesize $S=16$ slices};

\node[note] at (4.2,-3.0) {Poisson-to-rate annealing\\during training};
\node[note] at (12.2,-3.0) {Multi-instance learning\\at patient level};
\node[note] at (18.8,-3.0) {Inference averaged over\\multiple stochastic bags};

\end{tikzpicture}
}
\caption{Architecture of the ConvSNN-MIL model. Each patient is represented as a bag of axial slices. Every slice is processed by a shared convolutional spiking encoder, yielding slice-level feature vectors. These features are then aggregated by attention-based multiple-instance learning pooling to obtain a patient-level representation, which is mapped to the final lesion-presence probability.}
\label{fig:convsnn_mil_arch}
\end{figure}

\subsection{Network architectures}

Three spike-based model families were evaluated together with a conventional CNN baseline. The first spiking architecture was a convolutional spiking neural network (ConvSNN) trained with surrogate gradients. This model processes each preprocessed CT slice represented as a $96\times96\times15$ tensor obtained from 2.5D axial context and three HU windows. The input is processed by three convolutional leaky integrate-and-fire (LIF) blocks with channel progression $15\rightarrow24\rightarrow48\rightarrow96$, followed by global average pooling and a linear classifier, as illustrated in Fig.~\ref{fig:convsnn_arch}.

The second architecture was a multiple-instance learning extension of the convolutional SNN (ConvSNN-MIL). In this setting, each patient is represented as a bag of axial slices. Each slice is processed by the same convolutional spiking encoder, producing slice-level feature vectors. These features are then aggregated using attention-based pooling to obtain a patient-level representation, which is mapped to the final lesion-presence probability (Fig.~\ref{fig:convsnn_mil_arch}). During inference, predictions from multiple stochastic bags are averaged to reduce variance.

In addition, a Tempotron-style spike-timing classifier was evaluated as a compact timing-based reference model. Compared with the convolutional SNN variants, the Tempotron used a much simpler architecture and operated on time-to-first-spike encoded inputs.

The architectures of the convolutional SNN and ConvSNN-MIL models are illustrated in Fig.~\ref{fig:convsnn_arch} and Fig.~\ref{fig:convsnn_mil_arch}. The CNN baseline used the same convolutional backbone as the ConvSNN architecture but replaced spiking neurons with standard ReLU activations and removed temporal simulation. This ensured a fair comparison between spike-based and conventional dense computation.

\subsection{Surrogate Gradient Learning}

For differentiable training of spiking networks, surrogate gradient learning was employed \cite{zenke2021}. In the forward pass, spike generation was modeled as a hard threshold, while in the backward pass the derivative was approximated using the derivative of a logistic nonlinearity,
\begin{equation}
\frac{\partial S}{\partial V} \approx \sigma(a(V - V_{\mathrm{th}})) \big(1-\sigma(a(V - V_{\mathrm{th}}))\big), \qquad a>0.
\end{equation}
The convolutional SNN models were trained using a single-logit binary objective. In the ConvSNN-MIL setting, patient-level logits were obtained after attention-based aggregation of slice-level features. To address class imbalance, oversampling of positive training patients was applied, and in the MIL model an asymmetric focal loss was used to emphasize harder negative examples while preserving high sensitivity.

\subsection{Tempotron learning}

For spike-timing-based learning, a Tempotron-style classifier \cite{dong2022} was implemented using TTFS-encoded inputs. Given an input spike train $\mathbf{S}\in\{0,1\}^{T\times D}$, the membrane potential evolved as
\begin{equation}
v[t] = \alpha v[t-1] + \mathbf{w}^{\top}\mathbf{S}[t], \qquad \alpha = 0.95.
\end{equation}
A sample was classified as lesion-positive if the maximal membrane potential exceeded a decision threshold. During training, misclassified samples triggered weight updates proportional to the cumulative presynaptic activity up to the time of maximal membrane potential. This model served as a compact spike-timing baseline complementary to the surrogate-gradient convolutional SNNs.

\subsection{Patient-level aggregation and metrics}

All final evaluations were performed strictly at the patient level. For models producing slice-level predictions (CNN, ConvSNN, and Tempotron), all axial slices belonging to a given patient were evaluated independently during inference. Patient-level probabilities were obtained using a top-$1$ aggregation strategy, where the maximal slice-level probability within the patient volume was used as the final patient-level score. This aggregation rule reflects the clinical objective of detecting the most suspicious slice within a CT study and was selected based on validation performance. Because the endpoint was lesion presence rather than lesion burden or lesion count, top-$1$ aggregation was considered a clinically plausible patient-level decision surrogate.

For the ConvSNN-MIL architecture, patient-level predictions were obtained directly from attention-based pooling over slice features within each bag, and multiple stochastic bags per patient were averaged at inference time.

The patient-level decision threshold was selected on the validation set by maximizing the F1-score computed on patient-level predictions. This threshold was determined independently for each model and split and then transferred without modification to the corresponding held-out test patients. Performance was summarized using patient-level PR-AUC, ROC-AUC, F1-score, accuracy, balanced accuracy, specificity, sensitivity, Matthews correlation coefficient (MCC), Brier score, log-loss, and expected calibration error (ECE).

\subsection{Statistical analysis}

Performance metrics are reported as mean $\pm$ standard deviation across five independent patient-level splits. In addition, 95\% confidence intervals (CI) were estimated assuming approximate normality of the split-wise metric distributions.

Because all models were evaluated on identical patient-level splits, paired comparisons between models were treated as exploratory and interpreted cautiously due to the limited number of splits ($n=5$). Differences in performance were therefore primarily assessed based on effect sizes, consistency across splits, and metric stability rather than formal hypothesis testing alone.
Formal paired significance tests were deliberately not emphasized because the small number of repeated splits ($n=5$) limits both statistical power and the reliability of distributional assumptions required for inferential testing.

Because only five repeated patient-level splits were available, the estimated confidence intervals should be interpreted as approximate and primarily descriptive rather than inferential. No multiplicity-adjusted formal hypothesis testing was performed, and all between-model comparisons should therefore be interpreted as descriptive rather than confirmatory.

\subsection{Computational efficiency}

All experiments were executed on a single Intel Core i7-14700F CPU with 64 GB RAM. No GPU acceleration was used. For the CNN baseline and the convolutional SNN, computational cost was estimated using parameter counts together with MACs/FLOPs proxies derived from layer dimensions. For spiking models, we additionally estimated a SynOps-style proxy using the observed average spike rate during inference. In the ConvSNN-MIL setting, dense patient-level MACs/FLOPs and slice-level equivalents were also reported to characterize the computational footprint of bag-level inference.

\subsection{Implementation details and reproducibility}

All experiments were executed in a CPU-only setting using a fixed configuration file and five predefined random seeds $\{0,1,2,3,4\}$. To improve reproducibility, seed control was applied to Python, NumPy, and PyTorch random number generators before each run. The complete preprocessing and training setup was fixed across experiments, including image size ($96\times96$), bag size ($S=16$), axial context offsets $(-2,-1,0,1,2)$, three HU windows $[-100,200]$, $[0,300]$, and $[-150,250]$, patient-level label derivation from lesion masks, and liver ROI cropping. The Adam optimizer was used with an initial learning rate of $10^{-3}$ and default momentum parameters. Five repeated patient-level splits were selected as a practical compromise between variance estimation and the computational cost of CPU-only multi-model training across two datasets.

For all models, train/validation/test partitions were generated strictly at the patient level for each seed. Threshold tuning and aggregation choices were selected on the validation set and then transferred unchanged to the corresponding test set. Per-seed outputs included saved checkpoints, training histories, validation and test predictions, ROC and precision-recall curves, confusion matrices, and JSON summary files, enabling full result traceability and post hoc analysis.

All experiments were implemented in Python using PyTorch. The main software environment consisted of Python 3.10, PyTorch 2.2, NumPy 1.26, and scikit-learn 1.4. All experiments were executed on a CPU-only environment under Windows 11 Pro.

The entire experimental pipeline (preprocessing, training, evaluation, and metric aggregation) was executed using a fixed experiment script to ensure identical processing across all models and seeds. Detailed per-seed performance metrics, including patient-level predictions, thresholds, and confusion matrices, are provided in the supplementary material to ensure full reproducibility and facilitate independent verification of the reported results.

\subsection{Experimental transparency}

To ensure experimental transparency, all model architectures, preprocessing parameters, and evaluation procedures were fixed before running the five independent random splits. No model architecture or hyperparameter was modified after observing test set performance.

\subsection{Reproducibility statement}

To support reproducibility, all experiments were run using a fixed configuration file, predefined random seeds, and deterministic data partitioning at the patient level. The experimental outputs were stored systematically for each model and seed, including checkpoints, training histories, patient-level predictions, confusion matrices, ROC/PR plots, and aggregated result summaries. This setup enabled exact reconstruction of the reported metrics and facilitated consistent comparison across learning rules and architectures. Code supporting the findings of this study will be made available upon acceptance.

The main experimental configuration used in all experiments is summarized in Table~\ref{tab:config}.

\begin{table}[h]
\centering
\scriptsize
\caption{Main experimental configuration used across all models.}
\label{tab:config}
\begin{tabular}{lc}
\toprule
Parameter & Value \\
\midrule
Image size & $96 \times 96$ \\
Channels & 15 (2.5D + HU windows) \\
Bag size $S$ & 16 slices \\
Temporal steps $T$ & 8 \\
LIF decay $\beta$ & 0.95 \\
Spike threshold $V_{th}$ & 1.0 \\
Optimizer & Adam \\
Learning rate & $10^{-3}$ \\
Batch size & 8 patients \\
Training epochs & 12 \\
Random seeds & $\{0,1,2,3,4\}$ \\
Loss function & BCE (CNN, ConvSNN), focal loss (MIL) \\
Patient-level aggregation & top-1 (CNN, ConvSNN), attention MIL \\
Threshold selection & validation F1 maximization \\
Encoding & Poisson-to-rate (ConvSNN), TTFS (Tempotron) \\
Augmentation & flips, noise, gamma \\
\bottomrule
\end{tabular}
\end{table}

\section{Results}

We evaluated four models for binary liver lesion presence classification across two abdominal CT datasets under a strict leakage-free patient-level protocol: a conventional CNN baseline, a convolutional spiking neural network trained with surrogate gradients, an attention-based ConvSNN multiple-instance learning variant, and a Tempotron-style spike-timing classifier. All results were averaged over five independent patient-level splits. Because lesion-positive cases remained relatively limited at the test level, we report PR-AUC as the primary ranking metric, together with ROC-AUC, F1-score, accuracy, and MCC. Across the five random patient-level splits, the number of patients in the test set ranged between 25 and 27 depending on the stratified partitioning, with approximately balanced class proportions. This relatively small number of positive patients per split motivated reporting performance averaged across multiple seeds to obtain more stable estimates. Repeated evaluation across five independent patient-level splits improves robustness relative to a single train/validation/test partition and provides a more stable estimate of performance variability across patient-level partitions.

Table~\ref{tab:results_both_main} summarizes the main patient-level test results across both evaluated datasets. On \texttt{Task03\_Liver}, the CNN baseline achieved the highest mean patient-level PR-AUC ($0.970 \pm 0.036$), while the convolutional surrogate-gradient SNN showed the strongest threshold-dependent classification performance among the evaluated spiking models. In particular, the ConvSNN reached mean patient-level accuracy of $0.954 \pm 0.029$, F1-score of $0.957 \pm 0.027$, MCC of $0.913 \pm 0.054$, balanced accuracy of $0.954 \pm 0.029$, and ROC-AUC of $0.972 \pm 0.027$. Thus, although the CNN slightly outperformed all SNN models in PR-AUC, the ConvSNN yielded the best balance of discrimination and final binary decision quality.

The ConvSNN-MIL variant also performed strongly, with mean PR-AUC of $0.929 \pm 0.066$, ROC-AUC of $0.959 \pm 0.034$, accuracy of $0.915 \pm 0.045$, F1-score of $0.953 \pm 0.018$, and MCC of $0.880 \pm 0.018$. However, it did not improve upon the simpler ConvSNN. The Tempotron model produced the weakest results among the evaluated approaches, with mean PR-AUC of $0.905 \pm 0.038$, ROC-AUC of $0.886 \pm 0.053$, accuracy of $0.785 \pm 0.058$, F1-score of $0.779 \pm 0.065$, and MCC of $0.573 \pm 0.113$, indicating that timing-based spike learning alone was less effective on this task than convolutional surrogate-gradient optimization. In contrast, performance patterns differed on the Task03\_CECT dataset, where all models exhibited lower mean scores and higher variability across patient-level splits (Table~\ref{tab:results_both_main}). The CNN baseline retained the highest PR-AUC ($0.874 \pm 0.084$), but the performance gap between models was smaller than on \texttt{Task03\_Liver}. In particular, the ConvSNN-MIL model showed improved relative performance on Task03\_CECT (PR-AUC $0.804 \pm 0.124$, F1-score $0.766 \pm 0.124$, MCC $0.689 \pm 0.164$), outperforming the standard ConvSNN in several threshold-dependent metrics. The Tempotron model remained substantially weaker, confirming that timing-based spike learning alone did not generalize well across datasets.

These results indicate that model ranking is dataset-dependent and that conclusions drawn from a single benchmark may not generalize across CT cohorts. While convolutional SNN models remained competitive on both datasets, their relative advantage over the CNN baseline varied depending on the dataset characteristics and patient-level distribution.

\begin{table}[h!]
\centering
\scriptsize
\caption{Patient-level test performance across both evaluated CT datasets (mean $\pm$ std over 5 leakage-free patient-level splits). Primary ranking metric: PR-AUC.}
\label{tab:results_both_main}
\renewcommand{\arraystretch}{1.15}
\begin{tabular}{llcccc}
\toprule
\textbf{Dataset} & \textbf{Model} & \textbf{PR-AUC} & \textbf{ROC-AUC} & \textbf{F1} & \textbf{MCC} \\
\midrule

\multirow{4}{*}{\texttt{Task03\_Liver}}
& CNN baseline & $\mathbf{0.970 \pm 0.036}$ & $\mathbf{0.974 \pm 0.030}$ & $0.917 \pm 0.060$ & $0.836 \pm 0.114$ \\
& ConvSNN & $0.959 \pm 0.044$ & $0.972 \pm 0.027$ & $\mathbf{0.957 \pm 0.027}$ & $\mathbf{0.913 \pm 0.054}$ \\
& ConvSNN-MIL & $0.929 \pm 0.066$ & $0.959 \pm 0.034$ & $0.953 \pm 0.018$ & $0.880 \pm 0.018$ \\
& Tempotron & $0.905 \pm 0.038$ & $0.886 \pm 0.053$ & $0.779 \pm 0.065$ & $0.573 \pm 0.113$ \\

\midrule

\multirow{4}{*}{Task03\_CECT}
& CNN baseline & $\mathbf{0.874 \pm 0.084}$ & $\mathbf{0.952 \pm 0.035}$ & $\mathbf{0.802 \pm 0.099}$ & $\mathbf{0.733 \pm 0.134}$ \\
& ConvSNN & $0.766 \pm 0.119$ & $0.880 \pm 0.060$ & $0.708 \pm 0.112$ & $0.598 \pm 0.157$ \\
& ConvSNN-MIL & $0.804 \pm 0.124$ & $0.890 \pm 0.056$ & $0.766 \pm 0.124$ & $0.689 \pm 0.164$ \\
& Tempotron & $0.604 \pm 0.091$ & $0.790 \pm 0.045$ & $0.534 \pm 0.034$ & $0.368 \pm 0.069$ \\

\bottomrule
\end{tabular}
\end{table}

\begin{table}[h!]
\centering
\scriptsize
\caption{Additional patient-level classification metrics across both evaluated CT datasets (mean $\pm$ std over 5 leakage-free patient-level splits).}
\label{tab:results_both_additional}
\renewcommand{\arraystretch}{1.15}
\begin{tabular}{llcccc}
\toprule
\textbf{Dataset} & \textbf{Model} & \textbf{Accuracy} & \textbf{Balanced accuracy} & \textbf{Specificity} & \textbf{Sensitivity} \\
\midrule

\multirow{4}{*}{\texttt{Task03\_Liver}}
& CNN baseline & $0.915 \pm 0.057$ & $0.915 \pm 0.057$ & $0.877 \pm 0.038$ & $0.954 \pm 0.092$ \\
& ConvSNN & $\mathbf{0.954 \pm 0.029}$ & $\mathbf{0.954 \pm 0.029}$ & $\mathbf{0.908 \pm 0.058}$ & $\mathbf{1.000 \pm 0.000}$ \\
& ConvSNN-MIL & $0.915 \pm 0.045$ & $0.915 \pm 0.045$ & $0.831 \pm 0.090$ & $\mathbf{1.000 \pm 0.000}$ \\
& Tempotron & $0.785 \pm 0.058$ & $0.785 \pm 0.058$ & $0.800 \pm 0.062$ & $0.769 \pm 0.097$ \\

\midrule

\multirow{4}{*}{Task03\_CECT}
& CNN baseline & $\mathbf{0.896 \pm 0.054}$ & $\mathbf{0.868 \pm 0.069}$ & $\mathbf{0.926 \pm 0.043}$ & $0.811 \pm 0.109$ \\
& ConvSNN & $0.816 \pm 0.083$ & $0.821 \pm 0.078$ & $0.811 \pm 0.103$ & $\mathbf{0.832 \pm 0.114}$ \\
& ConvSNN-MIL & $0.879 \pm 0.063$ & $0.840 \pm 0.079$ & $0.922 \pm 0.056$ & $0.758 \pm 0.127$ \\
& Tempotron & $0.693 \pm 0.104$ & $0.687 \pm 0.034$ & $0.700 \pm 0.210$ & $0.674 \pm 0.215$ \\

\bottomrule
\end{tabular}
\end{table}

Additional patient-level classification metrics are reported in Table~\ref{tab:results_both_additional}. On \texttt{Task03\_Liver}, both convolutional SNN models achieved mean sensitivity of $1.000 \pm 0.000$ under the validation-selected operating thresholds used in this study, whereas the CNN baseline showed slightly lower sensitivity ($0.954 \pm 0.092$). At the same time, the ConvSNN achieved the highest mean specificity among the spiking models ($0.908 \pm 0.058$), indicating that the observed high recall was not obtained solely at the cost of excessive false positives. Because threshold selection was optimized on the validation set under a limited number of patient-level splits, this result should be interpreted as specific to the present operating-point configuration rather than as evidence of universally zero false-negative behavior.

On Task03\_CECT, the pattern was less pronounced. The CNN baseline achieved the highest mean accuracy, balanced accuracy, specificity, and sensitivity, while ConvSNN-MIL remained competitive and outperformed the standard ConvSNN in specificity. This suggests that the advantage of the convolutional spike-based models was less consistent on the second dataset and depended more strongly on the operating metric.

\begin{table}[h]
\centering
\scriptsize
\caption{Calibration metrics at the patient level across both evaluated CT datasets (mean $\pm$ std over 5 leakage-free patient-level splits). Lower values indicate better calibration.}
\label{tab:calibration_both}
\renewcommand{\arraystretch}{1.15}
\begin{tabular}{llcc}
\toprule
\textbf{Dataset} & \textbf{Model} & \textbf{ECE} $\downarrow$ & \textbf{Log-loss} $\downarrow$ \\
\midrule

\multirow{4}{*}{\texttt{Task03\_Liver}}
& CNN baseline & $0.431 \pm 0.092$ & $\mathbf{0.403 \pm 0.205}$ \\
& ConvSNN & $0.316 \pm 0.117$ & $0.679 \pm 0.015$ \\
& ConvSNN-MIL & $0.347 \pm 0.048$ & $0.698 \pm 0.028$ \\
& Tempotron & $\mathbf{0.248 \pm 0.078}$ & $0.648 \pm 0.082$ \\

\midrule

\multirow{4}{*}{Task03\_CECT}
& CNN baseline & $0.238 \pm 0.126$ & $0.498 \pm 0.224$ \\
& ConvSNN & $\mathbf{0.159 \pm 0.039}$ & $\mathbf{0.460 \pm 0.154}$ \\
& ConvSNN-MIL & $0.228 \pm 0.090$ & $0.499 \pm 0.045$ \\
& Tempotron & $0.543 \pm 0.050$ & $1.258 \pm 0.165$ \\

\bottomrule
\end{tabular}
\end{table}

Calibration results are summarized in Table~\ref{tab:calibration_both}. On \texttt{Task03\_Liver}, the CNN baseline achieved the lowest mean log-loss, indicating the best probabilistic calibration among the evaluated models, whereas the Tempotron achieved the lowest expected calibration error (ECE) despite substantially weaker classification performance. On Task03\_CECT, the ConvSNN achieved the most favorable calibration metrics, with the lowest mean ECE and log-loss, while the Tempotron showed clear miscalibration. These findings indicate that ranking quality, final decision quality, and probability calibration may differ across model families and datasets and should therefore be interpreted jointly.

Metric variability across random patient-level splits is summarized in Table~\ref{tab:seed_variability_both}. On \texttt{Task03\_Liver}, the convolutional surrogate-gradient SNN showed a favorable balance between strong mean performance and relatively low variability. Its standard deviations were lower than those of the CNN baseline for several threshold-dependent metrics, including MCC ($0.054$ \textit{versus}\ $0.114$), F1-score ($0.027$ \textit{versus}\ $0.060$), and accuracy ($0.029$ \textit{versus}\ $0.057$). At the same time, the ConvSNN-MIL model exhibited even lower variability for selected threshold-dependent metrics, indicating that stability and absolute performance did not fully coincide across model families.

On Task03\_CECT, variability patterns were different. The CNN baseline showed the lowest standard deviations for F1-score, MCC, and accuracy, whereas the Tempotron exhibited low variability for F1-score and MCC but at substantially lower absolute performance. Thus, stability and mean predictive quality did not coincide uniformly across datasets.

\begin{table}[h]
\centering
\scriptsize
\caption{Metric variability across five independent seeds for both evaluated CT datasets. Lower values indicate more stable performance across patient-level splits.}
\label{tab:seed_variability_both}
\begin{tabular}{llccc}
\toprule
\textbf{Dataset} & \textbf{Model} & \textbf{F1 std} & \textbf{MCC std} & \textbf{Accuracy std} \\
\midrule

\multirow{4}{*}{\texttt{Task03\_Liver}}
& CNN & 0.060 & 0.114 & 0.057 \\
& ConvSNN & \textbf{0.027} & \textbf{0.054} & \textbf{0.029} \\
& ConvSNN-MIL & 0.018 & 0.018 & 0.045 \\
& Tempotron & 0.065 & 0.113 & 0.058 \\

\midrule

\multirow{4}{*}{Task03\_CECT}
& CNN & \textbf{0.099} & \textbf{0.134} & \textbf{0.054} \\
& ConvSNN & 0.112 & 0.157 & 0.083 \\
& ConvSNN-MIL & 0.124 & 0.164 & 0.063 \\
& Tempotron & 0.034 & 0.069 & 0.104 \\

\bottomrule
\end{tabular}
\end{table}

\begin{table}[h]
\centering
\scriptsize
\caption{Approximate computational complexity of evaluated models. 
For CNN and ConvSNN, MACs/FLOPs are reported per slice. 
For ConvSNN-MIL, the dense cost reflects bag-level inference and is therefore reported separately in the text. 
For spiking models, SynOps denotes a spike-rate-adjusted proxy of effective synaptic operations.}
\label{tab:compute}
\begin{tabular}{lcccc}
\toprule
Model & Parameters & MACs/slice & FLOPs/slice & SynOps proxy \\
\midrule
CNN & $1.14\times10^5$ & $7.75\times10^7$ & $1.55\times10^8$ & -- \\
ConvSNN & $1.13\times10^5$ & $7.75\times10^7$ & $1.55\times10^8$ & $5.82\times10^6$ \\
ConvSNN-MIL & -- & $1.55\times10^9$ & $3.11\times10^9$ & $1.68\times10^8$ \\
Tempotron & $2.63\times10^2$ & $2.63\times10^2$ & $5.26\times10^2$ & -- \\
\bottomrule
\end{tabular}
\end{table}

\begin{table}[h]
\centering
\tiny
\caption{Validation-selected decision thresholds averaged over five seeds for both evaluated CT datasets.}
\label{tab:thresholds_both}
\begin{tabular}{llc}
\toprule
\textbf{Dataset} & \textbf{Model} & \textbf{Mean threshold} \\
\midrule

\multirow{4}{*}{\texttt{Task03\_Liver}}
& CNN & 0.717 \\
& ConvSNN & 0.641 \\
& ConvSNN-MIL & 0.680 \\
& Tempotron & 0.534 \\

\midrule

\multirow{4}{*}{Task03\_CECT}
& CNN & 0.480 \\
& ConvSNN & 0.235 \\
& ConvSNN-MIL & 0.355 \\
& Tempotron & 0.815 \\

\bottomrule
\end{tabular}
\end{table}

\begin{table}[h!]
\centering
\tiny
\caption{Cross-dataset comparison of the evaluated models. Results are shown as mean performance over 5 leakage-free patient-level splits.}
\label{tab:cross_dataset_summary}
\renewcommand{\arraystretch}{1.15}
\begin{tabular}{lcccccc}
\toprule
\multirow{2}{*}{\textbf{Model}} & \multicolumn{2}{c}{\textbf{PR-AUC}} & \multicolumn{2}{c}{\textbf{F1}} & \multicolumn{2}{c}{\textbf{MCC}} \\
\cmidrule(lr){2-3}\cmidrule(lr){4-5}\cmidrule(lr){6-7}
& \textbf{\texttt{Task03\_Liver}} & \textbf{Task03\_CECT} & \textbf{\texttt{Task03\_Liver}} & \textbf{Task03\_CECT} & \textbf{\texttt{Task03\_Liver}} & \textbf{Task03\_CECT} \\
\midrule
CNN baseline & 0.970 & 0.874 & 0.917 & 0.802 & 0.836 & 0.733 \\
ConvSNN & 0.959 & 0.766 & 0.957 & 0.708 & 0.913 & 0.598 \\
ConvSNN-MIL & 0.929 & 0.804 & 0.953 & 0.766 & 0.880 & 0.689 \\
Tempotron & 0.905 & 0.604 & 0.779 & 0.534 & 0.573 & 0.368 \\
\bottomrule
\end{tabular}
\end{table}

Table~\ref{tab:cross_dataset_summary} highlights that while absolute performance decreased on Task03\_CECT, the relative ordering of models was not fully consistent across datasets, indicating sensitivity to data distribution and emphasizing the need for multi-dataset evaluation in medical imaging benchmarks.

Computational characteristics are summarized in Table~\ref{tab:compute}. From a dense-compute perspective, the CNN baseline and the ConvSNN had nearly identical architectural complexity, with approximately $1.14\times10^{5}$ and $1.13\times10^{5}$ trainable parameters, respectively, and identical per-slice MAC/FLOP proxies ($7.75\times10^{7}$ MACs and $1.55\times10^{8}$ FLOPs). However, the ConvSNN additionally exhibited sparse event activity, with a mean test spike rate of $0.0038 \pm 0.0009$, corresponding to a SynOps-style proxy of $5.82\times10^{6} \pm 1.35\times10^{6}$ effective operations. The ConvSNN-MIL model was substantially heavier because of bag-level inference, with dense patient-level complexity of $2.49\times10^{10}$ MACs and $4.97\times10^{10}$ FLOPs per patient, but still maintained a low mean spike rate of $0.0068 \pm 0.0010$ and a SynOps-style proxy of $1.68\times10^{8} \pm 2.57\times10^{7}$. These findings suggest that spike-based models can achieve competitive predictive performance while maintaining sparse internal activity, which may be advantageous for future event-driven or neuromorphic deployment. However, these potential efficiency benefits were not directly validated on neuromorphic hardware in the present study.

Finally, Table~\ref{tab:thresholds_both} reports the validation-selected decision thresholds averaged over seeds. On \texttt{Task03\_Liver}, the CNN required the highest average threshold, the ConvSNN operated at an intermediate threshold, and the Tempotron used the lowest one, with ConvSNN-MIL again lying between the ConvSNN and CNN settings. On Task03\_CECT, the pattern differed, with the Tempotron requiring the highest mean threshold and the ConvSNN the lowest, further indicating that threshold transfer and score calibration were dataset-dependent.

Detailed per-seed patient-level results, approximate confidence intervals, and cross-seed metric ranges are provided in the Supplementary Material (Tables~\ref{tab:per_seed_full}--\ref{tab:spread_metrics}). Across the five leakage-free patient-level splits on \texttt{Task03\_Liver}, the CNN baseline showed a wider spread for threshold-dependent metrics than the standard ConvSNN. In particular, the CNN F1-score ranged from 0.800 to 0.963 and MCC from 0.617 to 0.926, whereas the ConvSNN ranged from 0.929 to 1.000 for F1-score and from 0.856 to 1.000 for MCC. The ConvSNN-MIL model remained highly stable for F1-score, MCC, and accuracy, but displayed a wider PR-AUC range (0.840--1.000), indicating that ranking stability and threshold-dependent decision stability do not necessarily coincide.

\begin{figure*}[t]
\centering

\begin{subfigure}{0.48\textwidth}
\centering
\includegraphics[width=\linewidth]{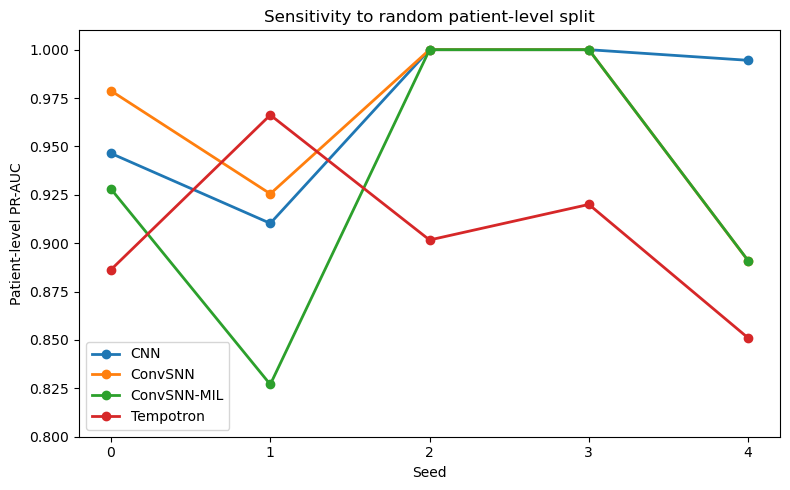}
\caption{\texttt{Task03\_Liver}}
\label{fig:seed_pr_auc_msd}
\end{subfigure}
\hfill
\begin{subfigure}{0.48\textwidth}
\centering
\includegraphics[width=\linewidth]{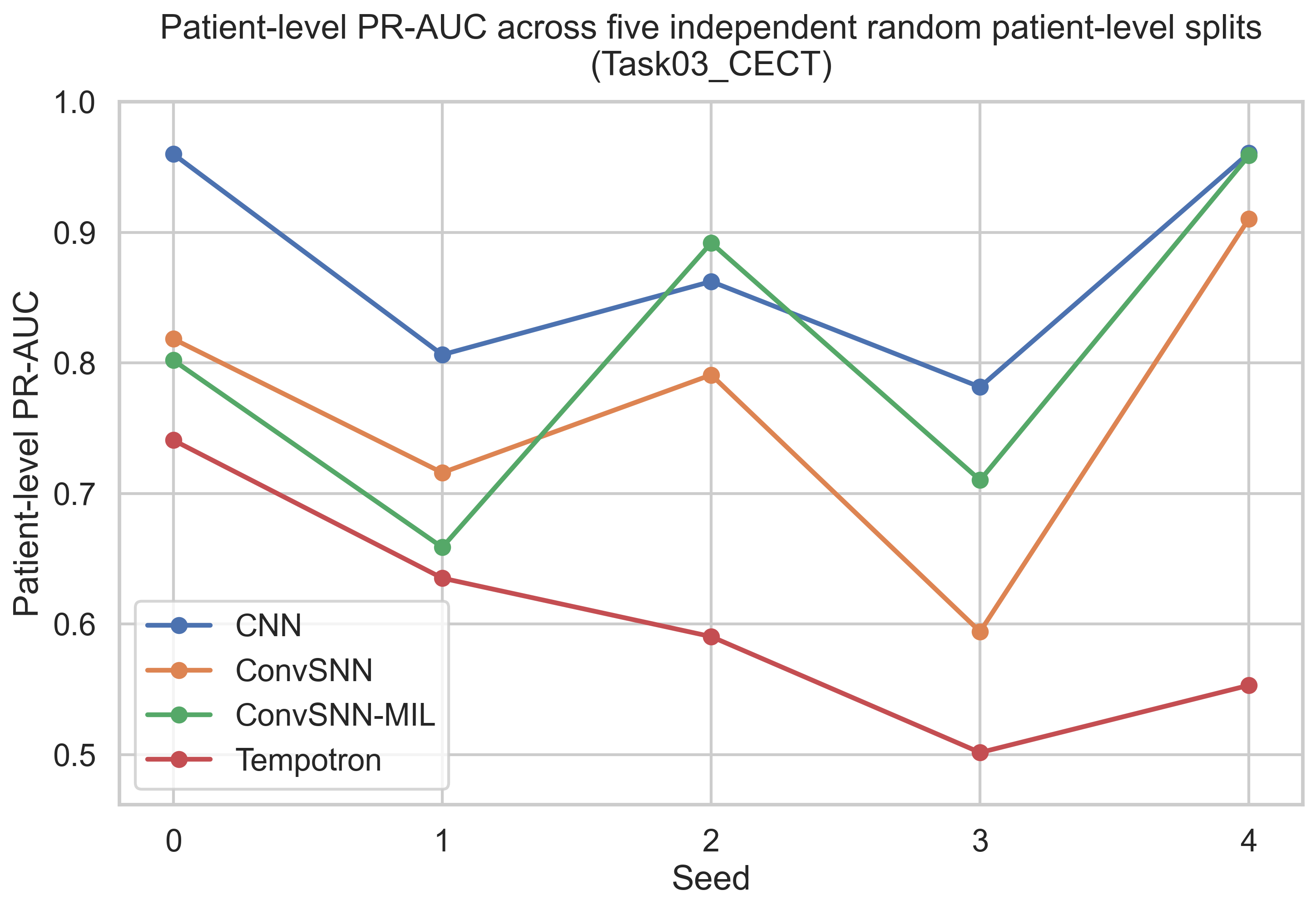}
\caption{\texttt{Task03\_CECT}}
\label{fig:seed_pr_auc_cect}
\end{subfigure}

\caption{
Seed-wise patient-level PR-AUC across five independent leakage-free patient-level splits for the two evaluated CT datasets.
Panel (a) shows results for \texttt{Task03\_Liver}, and panel (b) for \texttt{Task03\_CECT}. 
The second dataset exhibited lower absolute PR-AUC and greater variability across seeds, indicating stronger sensitivity to the particular patient-level partition.
}
\label{fig:seed_pr_auc_both}
\end{figure*}

\begin{figure*}[t]
\centering

\begin{subfigure}{0.24\textwidth}
\centering
\includegraphics[width=\linewidth]{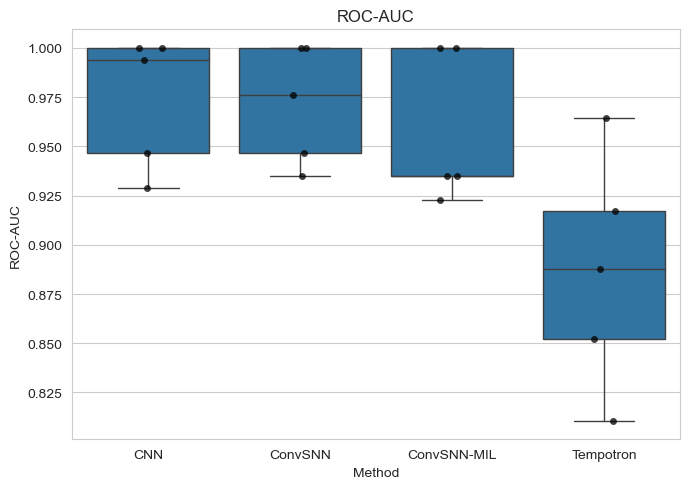}
\caption{MSD: PR-AUC}
\end{subfigure}
\hfill
\begin{subfigure}{0.24\textwidth}
\centering
\includegraphics[width=\linewidth]{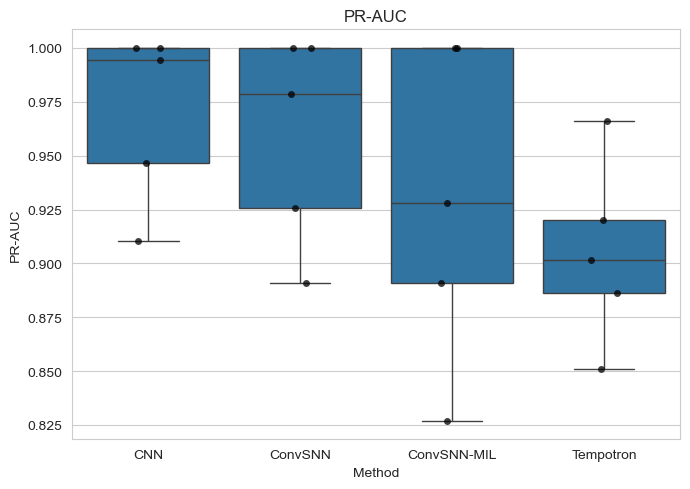}
\caption{MSD: ROC-AUC}
\end{subfigure}
\hfill
\begin{subfigure}{0.24\textwidth}
\centering
\includegraphics[width=\linewidth]{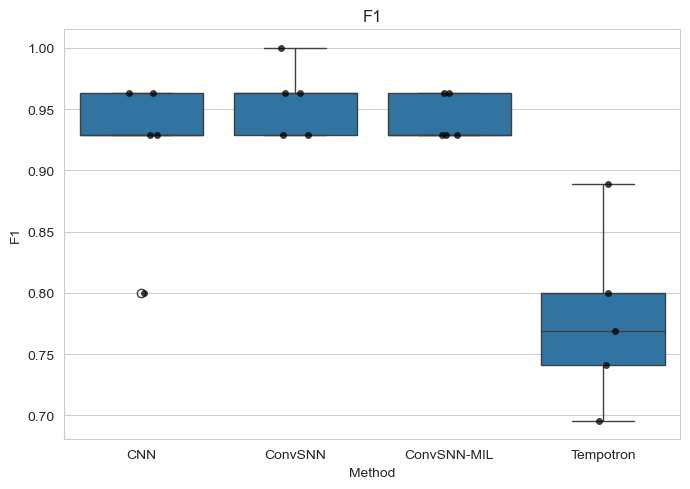}
\caption{MSD: F1-score}
\end{subfigure}
\hfill
\begin{subfigure}{0.24\textwidth}
\centering
\includegraphics[width=\linewidth]{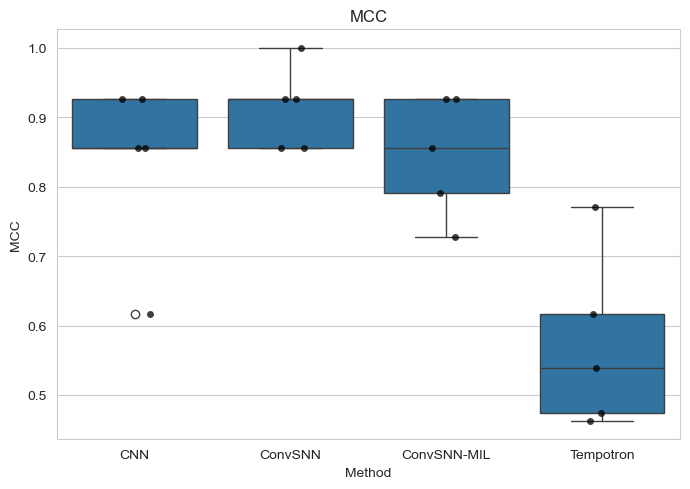}
\caption{MSD: MCC}
\end{subfigure}

\vspace{4mm}

\begin{subfigure}{0.24\textwidth}
\centering
\includegraphics[width=\linewidth]{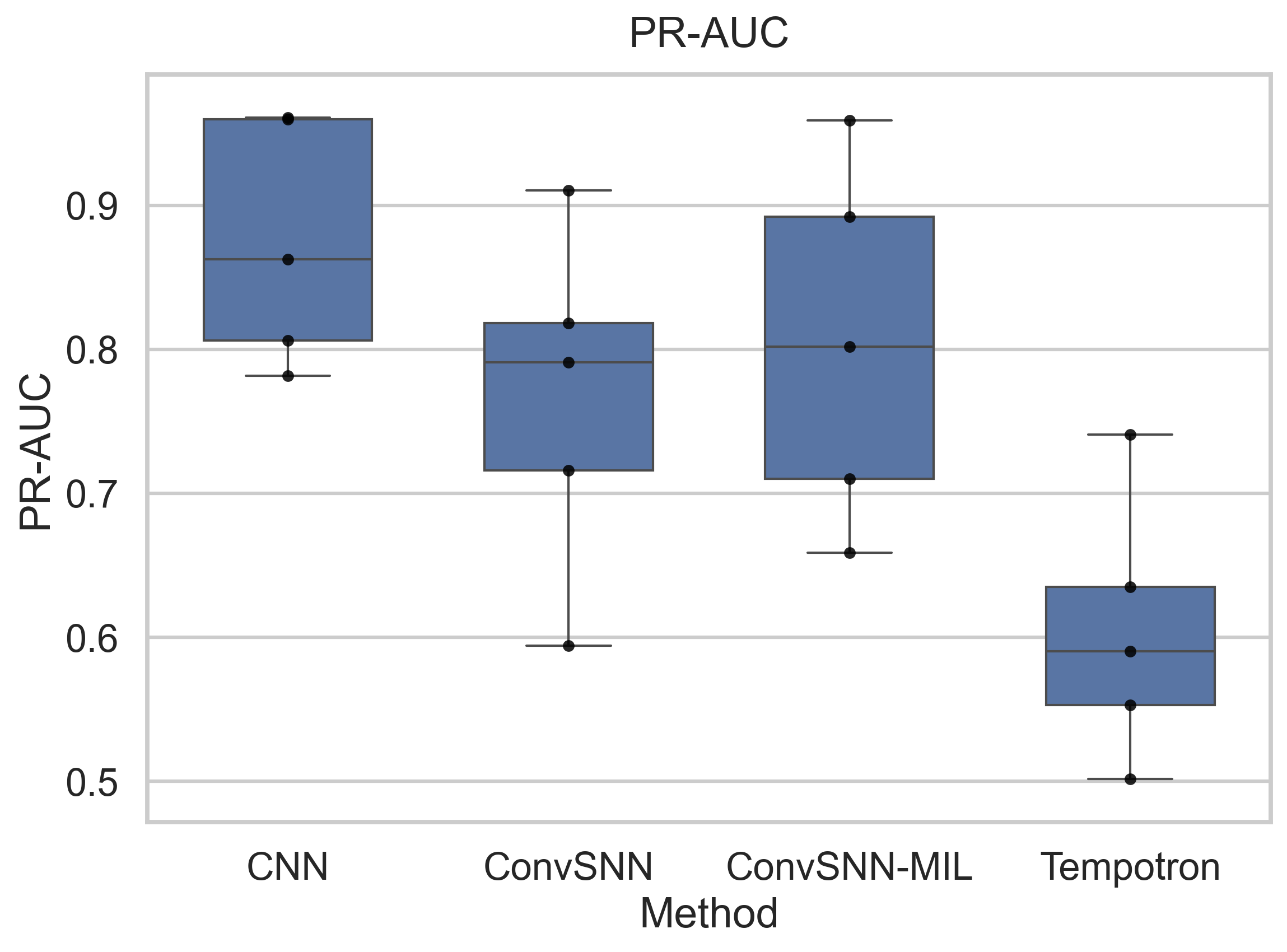}
\caption{CECT: PR-AUC}
\end{subfigure}
\hfill
\begin{subfigure}{0.24\textwidth}
\centering
\includegraphics[width=\linewidth]{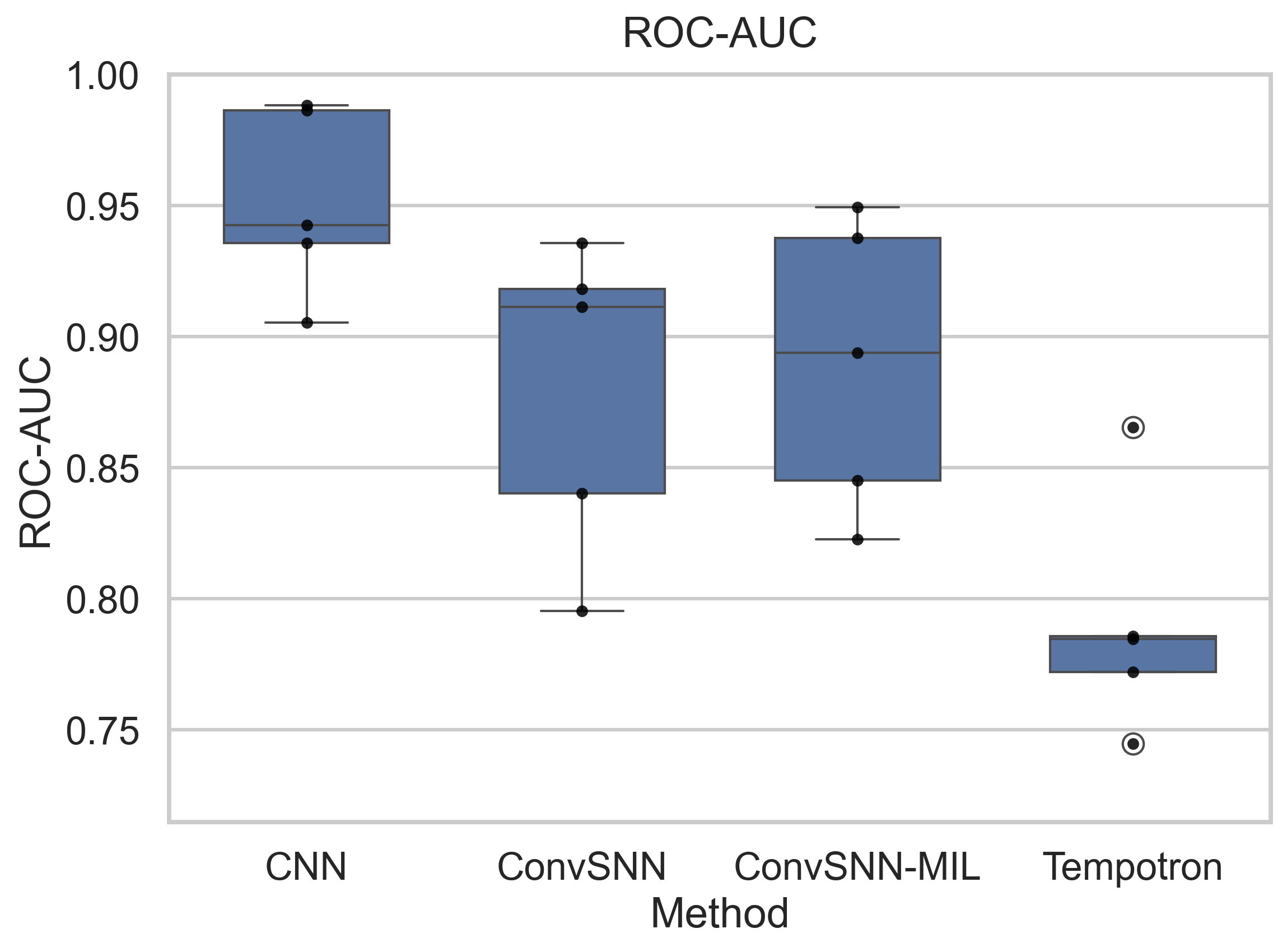}
\caption{CECT: ROC-AUC}
\end{subfigure}
\hfill
\begin{subfigure}{0.24\textwidth}
\centering
\includegraphics[width=\linewidth]{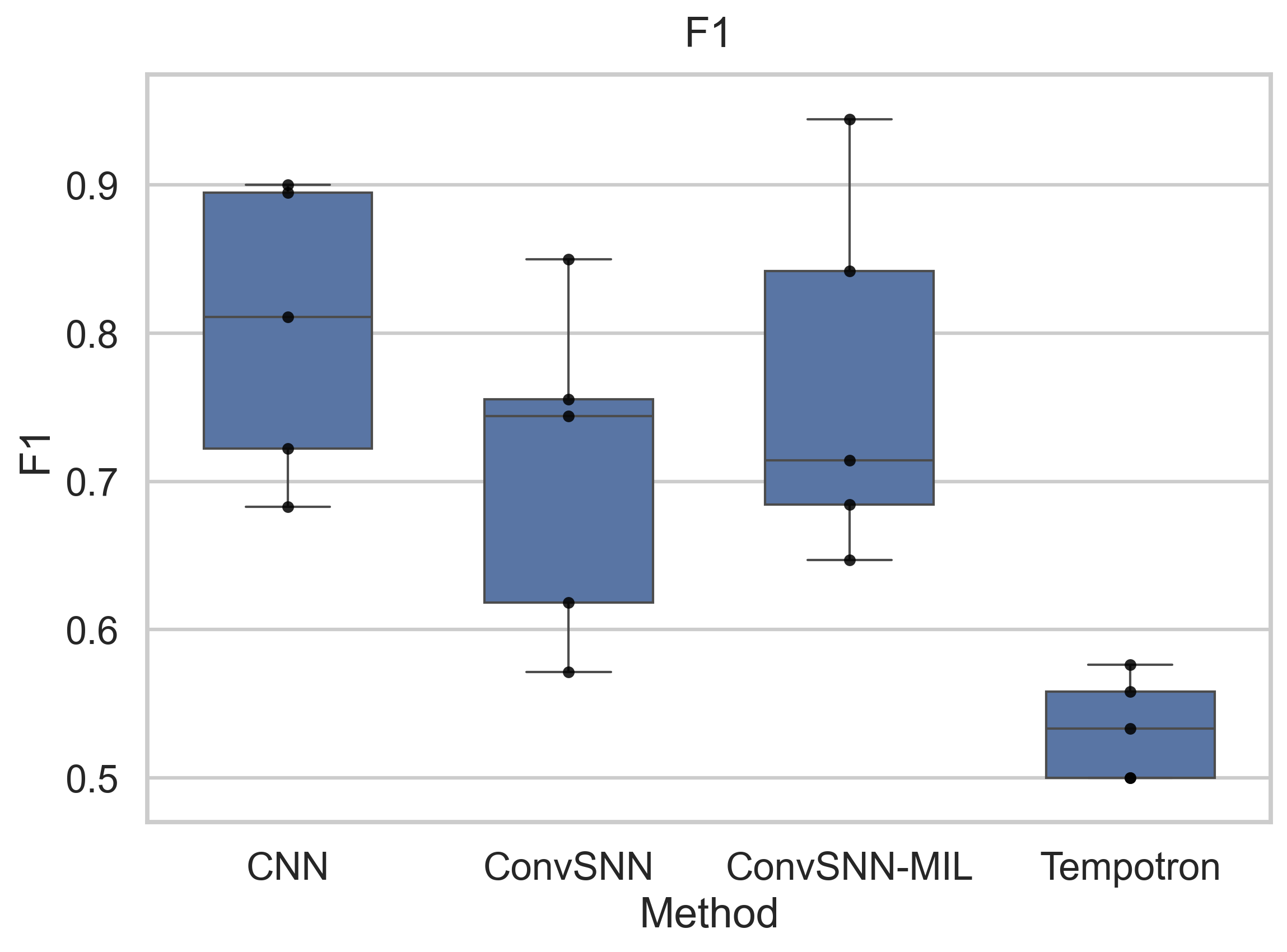}
\caption{CECT: F1-score}
\end{subfigure}
\hfill
\begin{subfigure}{0.24\textwidth}
\centering
\includegraphics[width=\linewidth]{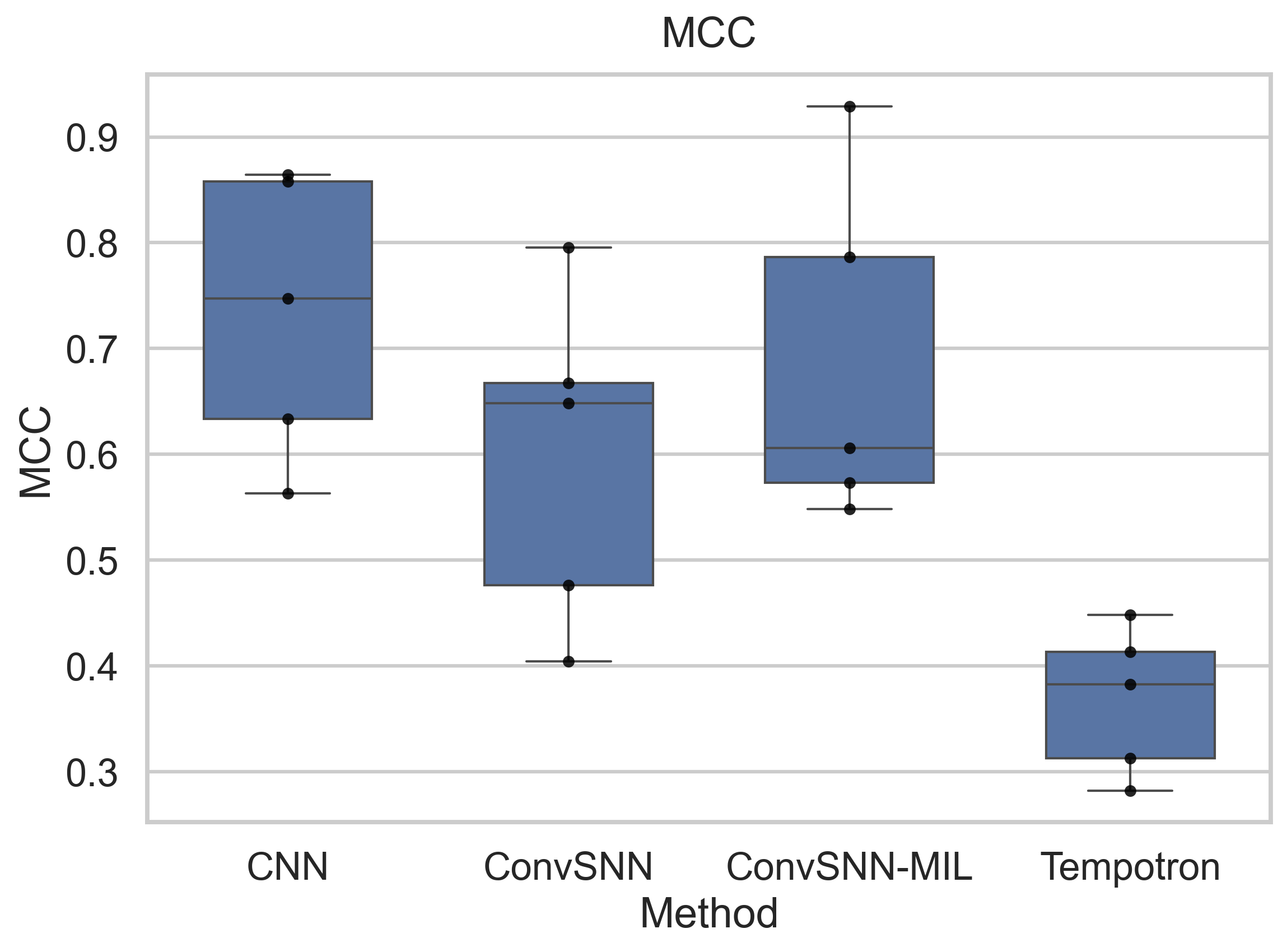}
\caption{CECT: MCC}
\end{subfigure}

\caption{
Distribution of patient-level performance across five independent random patient-level splits for both evaluated datasets.
The top row shows \texttt{Task03\_Liver}, and the bottom row shows \texttt{Task03\_CECT}. 
Across the main metrics, \texttt{Task03\_Liver} exhibited higher and more stable performance, whereas \texttt{Task03\_CECT} showed a broader performance spread and greater dataset-dependent variability.
}
\label{fig:boxplots_both}
\end{figure*}

Figure~\ref{fig:seed_pr_auc_both} shows the seed-wise variation in patient-level PR-AUC across the two evaluated datasets. In panel (a) of Figure~\ref{fig:seed_pr_auc_both}, the CNN and ConvSNN maintained consistently high PR-AUC across the five independent patient-level splits on \texttt{Task03\_Liver}, whereas the Tempotron exhibited lower and more variable ranking performance. In panel (b) of Figure~\ref{fig:seed_pr_auc_both}, all models showed greater seed-to-seed variability and lower absolute PR-AUC values on \texttt{Task03\_CECT}, confirming that the second dataset posed a more challenging and less stable evaluation setting.

Figure~\ref{fig:boxplots_both} summarizes the cross-seed distributions of PR-AUC, ROC-AUC, F1-score, and MCC for both datasets. The top row of Figure~\ref{fig:boxplots_both} indicates that \texttt{Task03\_Liver} yielded consistently strong and relatively stable performance across models, particularly for the CNN and ConvSNN. In contrast, the bottom row shows a broader performance spread on \texttt{Task03\_CECT}, with reduced median performance and increased variability across patient-level splits. Together, these visualizations support the conclusion that model ranking and apparent robustness were more dataset-dependent on the second benchmark.

\section{Ablation Study}

To better understand which design choices were most important for the final performance, we conducted targeted ablation analyses focused on the convolutional surrogate-gradient SNN pipeline across both evaluated CT datasets. In particular, we examined the role of convolutional feature extraction and the effect of patient-level aggregation strategy under the same leakage-free patient-level evaluation protocol.

First, we compared the convolutional surrogate-gradient SNN with the simpler Tempotron-style spike-timing classifier. As shown in Tables~\ref{tab:ablation_conv_msd} and \ref{tab:ablation_conv_cect}, the convolutional spiking model substantially outperformed the timing-only baseline across all main patient-level metrics on both datasets. This indicates that explicit convolutional feature extraction is essential for capturing the complex spatial structure of abdominal CT data, whereas spike timing alone is insufficient for this task.

Second, we compared two patient-level decision formulations: top-$1$ aggregation of slice-level predictions in the standard ConvSNN pipeline and attention-based multiple-instance learning in ConvSNN-MIL. The results are summarized in Tables~\ref{tab:ablation_agg_msd} and \ref{tab:ablation_agg_cect}. On \texttt{Task03\_Liver}, the standard ConvSNN with top-$1$ aggregation achieved higher PR-AUC, F1-score, and MCC than the MIL-based variant. In contrast, on Task03\_CECT, the attention-based ConvSNN-MIL variant achieved higher PR-AUC, F1-score, MCC, and accuracy than the standard top-$1$ aggregation strategy. This suggests that the relative benefit of top-$1$ slice aggregation versus learned attention-based bag aggregation was dataset-dependent rather than universal for the present binary lesion-presence task.

For the slice-based models, retrospective comparison of alternative aggregation rules such as top-$k$ and mean aggregation was not informative on the final exported test files, because these files stored only one final patient-level probability per case after the validation-selected aggregation step. Therefore, the final ablation analysis focuses on the aggregation paradigm used in the deployed pipelines rather than on recomputing multiple post hoc aggregation rules from already aggregated outputs.

Overall, the ablation results indicate that the strongest gains arose from convolutional spike-based feature extraction rather than from spike timing alone. In contrast, the optimal patient-level decision strategy depended on the dataset: top-$1$ slice aggregation was more effective on \texttt{Task03\_Liver}, whereas attention-based MIL pooling was more effective on Task03\_CECT.

\begin{table}[h]
\centering
\footnotesize
\caption{Ablation of convolutional feature extraction on \texttt{Task03\_Liver}.}
\label{tab:ablation_conv_msd}
\begin{tabular}{lcccc}
\toprule
Model & PR-AUC & ROC-AUC & F1 & MCC \\
\midrule
Tempotron (timing only) & $0.905 \pm 0.038$ & $0.886 \pm 0.053$ & $0.779 \pm 0.065$ & $0.573 \pm 0.113$ \\
ConvSNN (convolutional SNN) & $\mathbf{0.959 \pm 0.044}$ & $\mathbf{0.972 \pm 0.027}$ & $\mathbf{0.957 \pm 0.027}$ & $\mathbf{0.913 \pm 0.054}$ \\
\bottomrule
\end{tabular}
\end{table}

\begin{table}[h]
\centering
\footnotesize
\caption{Ablation of convolutional feature extraction on Task03\_CECT.}
\label{tab:ablation_conv_cect}
\begin{tabular}{lcccc}
\toprule
Model & PR-AUC & ROC-AUC & F1 & MCC \\
\midrule
Tempotron (timing only) & $0.604 \pm 0.091$ & $0.790 \pm 0.045$ & $0.534 \pm 0.034$ & $0.368 \pm 0.069$ \\
ConvSNN (convolutional SNN) & $\mathbf{0.766 \pm 0.119}$ & $\mathbf{0.880 \pm 0.060}$ & $\mathbf{0.708 \pm 0.112}$ & $\mathbf{0.598 \pm 0.157}$ \\
\bottomrule
\end{tabular}
\end{table}

\begin{table}[h]
\centering
\footnotesize
\caption{Comparison of patient-level decision strategies on \texttt{Task03\_Liver}: slice-based top-$1$ aggregation versus attention-based MIL pooling.}
\label{tab:ablation_agg_msd}
\begin{tabular}{lcccc}
\toprule
Model / strategy & PR-AUC & F1 & MCC & Accuracy \\
\midrule
ConvSNN + top-$1$ slice aggregation & $\mathbf{0.959 \pm 0.044}$ & $\mathbf{0.957 \pm 0.027}$ & $\mathbf{0.913 \pm 0.054}$ & $\mathbf{0.954 \pm 0.029}$ \\
ConvSNN-MIL + attention pooling & $0.929 \pm 0.066$ & $0.953 \pm 0.018$ & $0.880 \pm 0.018$ & $0.915 \pm 0.045$ \\
\bottomrule
\end{tabular}
\end{table}

\begin{table}[h]
\centering
\footnotesize
\caption{Comparison of patient-level decision strategies on Task03\_CECT: slice-based top-$1$ aggregation versus attention-based MIL pooling.}
\label{tab:ablation_agg_cect}
\begin{tabular}{lcccc}
\toprule
Model / strategy & PR-AUC & F1 & MCC & Accuracy \\
\midrule
ConvSNN + top-$1$ slice aggregation & $0.766 \pm 0.119$ & $0.708 \pm 0.112$ & $0.598 \pm 0.157$ & $0.816 \pm 0.083$ \\
ConvSNN-MIL + attention pooling & $\mathbf{0.804 \pm 0.124}$ & $\mathbf{0.766 \pm 0.124}$ & $\mathbf{0.689 \pm 0.164}$ & $\mathbf{0.879 \pm 0.063}$ \\
\bottomrule
\end{tabular}
\end{table}

These observations are consistent with the main results section and indicate that the strongest architectural gains arose from convolutional spike-based feature extraction. However, the optimal patient-level decision strategy was dataset-dependent: top-$1$ slice aggregation was more effective on \texttt{Task03\_Liver}, whereas attention-based MIL pooling was more effective on Task03\_CECT.

\section{Discussion}

A critical methodological aspect of this study is the use of strictly patient-level data splitting and evaluation across two abdominal CT datasets. This design prevents information leakage across highly correlated slices from the same CT volume and ensures that reported results reflect true case-level generalization. While direct quantitative comparison with prior studies is challenging because of differences in datasets, modalities, and clinical endpoints, the proposed SNNDeep framework demonstrates competitive performance under a leakage-free patient-level protocol that is still rarely adopted in existing SNN-based medical imaging studies.

To obtain robust estimates under patient-level class imbalance, all experiments were repeated across five independent random patient-level splits. For each split, aggregation choices and decision thresholds were selected on the validation patients and then transferred unchanged to the held-out test patients. In the final configuration, top-$1$ aggregation was used for the slice-based models, whereas attention-based MIL pooling provided an alternative patient-level formulation. The ablation results further showed that the relative benefit of these aggregation strategies was dataset-dependent rather than universal.

An important additional finding of this study is that model ranking was dataset-dependent. On \texttt{Task03\_Liver}, the CNN baseline achieved the highest mean patient-level PR-AUC, whereas the convolutional surrogate-gradient SNN showed the strongest threshold-dependent classification performance among the evaluated spiking models. On Task03\_CECT, absolute performance decreased for all models and variability across random patient-level splits increased, indicating that the second dataset posed a more challenging benchmark. These observations suggest that conclusions about the utility of spike-based models in medical imaging should not be drawn from a single dataset alone.

While part of the observed performance may be related to the high spatial resolution and reduced operator dependence of CT, the consistent behavior of SNNDeep across the evaluated learning paradigms suggests that architectural and algorithmic factors also play an important role. The network’s temporal encoding, biologically inspired learning mechanisms, and fine-grained hyperparameter tuning suggest that the improvements are not modality-specific but stem from the model’s ability to exploit temporal structure and spike dynamics.

Comparison with the metaheuristic-trained SNN proposed by \cite{kaur2021}, evaluated on the UCI Liver Disorders dataset, should be interpreted cautiously because of substantial differences in dataset characteristics, feature space, and clinical endpoint. Nevertheless, the present results suggest that convolutional surrogate-gradient SNNs can remain competitive under a stricter patient-level imaging benchmark. Because lesion-positive cases are rare at the patient level, accuracy alone is not a reliable indicator of clinical usefulness. In several splits, models achieved high ROC-AUC while exhibiting large variations in F1 and MCC due to threshold transfer from validation to test and the small number of positive test patients. Therefore, we emphasize patient-level PR-AUC and multi-seed reporting, which provide a more appropriate assessment of performance under strong imbalance. Under this protocol, the CNN baseline achieved the highest mean patient-level PR-AUC, indicating the strongest overall ranking performance across thresholds. However, among the evaluated spiking models, the convolutional surrogate-gradient SNN achieved competitive performance relative to the CNN baseline, particularly for threshold-dependent metrics such as accuracy, F1-score, and MCC. This distinction is important: PR-AUC reflects ranking quality, whereas the threshold-dependent metrics better capture final binary decision quality under clinically relevant operating points.

Direct comparison with previously published studies should be interpreted with caution, since many reported liver imaging models address different clinical endpoints, imaging modalities, or evaluation units. In particular, prior work has often focused on fibrosis staging, lesion characterization, or reconstruction tasks rather than binary patient-level lesion-presence classification. For this reason, the present results are best interpreted primarily as an internally consistent benchmark of SNN and CNN models under a leakage-free patient-level protocol rather than as a direct performance comparison with heterogeneous literature reports.

We observed substantial variability across random patient-level splits, reflected by wide confidence intervals for PR-AUC in some seeds and by the broader spread of several metrics on Task03\_CECT. This behavior is expected in the presence of severe patient-level imbalance and a limited number of lesion-positive patients per test split. Consequently, we report mean $\pm$ standard deviation over five seeds and interpret single-split results with caution. This variability further supports the use of repeated patient-level evaluation rather than reliance on a single train/validation/test partition, which could otherwise yield overly optimistic or unstable conclusions.

Beyond SNN-based methods, conventional convolutional neural networks have also been extensively applied to the classification and staging of liver disease. Another study \cite{choi2018} developed a deep CNN to stage liver fibrosis using contrast-enhanced CT, achieving 79.40\% precision with AUCs of 0.95--0.97 for different thresholds of fibrosis. Another study \cite{yasaka2018} applied a deep CNN to gadoxetic acid-enhanced hepatobiliary-phase MRI, obtaining AUCs of approximately 0.84--0.85 for the detection of advanced fibrosis. More recently, \cite{yin2021} used a deep CNN with Grad-CAM explainability to stage liver fibrosis on MRI, reporting AUC values of 0.88--0.92. While these CNN-based models demonstrate clinical feasibility and good performance, direct numerical comparison with the present study remains difficult because of differences in modality, endpoint definition, and evaluation protocol. Furthermore, dense convolutional operations increase computational cost, making them less adaptable to event-driven or low-power neuromorphic deployment.

The Tempotron was included as a compact timing-based reference model rather than as an architecture matched in representational capacity to the convolutional SNN variants. Within the evaluated spiking configurations, the convolutional surrogate-gradient SNN achieved the strongest balance of threshold-dependent metrics on the primary MSD benchmark, including mean patient-level accuracy above 95\%, together with high ROC-AUC and MCC. The ConvSNN-MIL model remained competitive across both datasets, although its relative advantage over the simpler convolutional SNN was dataset-dependent. In contrast, the Tempotron classifier was consistently weaker, supporting the conclusion that the strongest gains arose not from spike timing alone, but from the combination of convolutional feature extraction and surrogate-gradient optimization. From a computational perspective, this result highlights the importance of spatial feature extraction prior to spike-based decision mechanisms. While spike timing encodes temporal structure, convolutional layers provide translation-invariant spatial representations that capture anatomical patterns in CT images. The combination of convolutional feature hierarchies with spike-based neuronal dynamics therefore appears to provide a complementary mechanism: convolutional layers extract spatially structured features, while spiking neurons perform temporally integrated evidence accumulation.

From a translational perspective, the proposed framework aligns with practical considerations for deployment in clinical environments \cite{rudnicka2026}. Its compact three-layer design and compatibility with event-driven computation make it suitable for energy-efficient neuromorphic hardware (e.g., Intel Loihi, SpiNNaker) or edge-computing PACS integration. Such deployment could facilitate real-time decision support in radiology workflows, particularly in settings with limited access to subspecialist expertise. Prospective validation on independent, multicenter cohorts will be a critical step toward clinical translation, ensuring robustness across different scanner protocols, institutions, and patient populations. Integration with automated liver segmentation pipelines and PACS systems will be essential for practical clinical deployment.

An important observation is that dense complexity alone does not fully characterize the computational profile of spiking models. In our experiments, the CNN and ConvSNN had almost identical parameter counts and dense MAC/FLOP proxies, which means that the architectural comparison was not biased in favor of the SNN by using a substantially larger conventional baseline. At the same time, the ConvSNN operated with a very low average spike rate, leading to a much smaller SynOps-style proxy than its dense equivalent. This indicates that the model maintained strong classification performance while relying on sparse internal activity, a property that is particularly relevant for future deployment on neuromorphic or event-driven hardware. However, these potential efficiency benefits were not directly validated on neuromorphic hardware in this study. Therefore, the reported SynOps-style estimates should be interpreted as architecture-level efficiency proxies rather than direct measurements of deployment-time energy consumption or wall-clock acceleration. Across the evaluated architectures, model complexity varied substantially. The number of trainable parameters ranged from $2.63\times10^2$ for the Tempotron to approximately $1.14\times10^5$ for the convolutional models, while slice-level MAC estimates ranged from $2.63\times10^2$ to $1.55\times10^9$. This highlights a clear trade-off between architectural expressiveness and computational cost, while also showing that sparse spike-based activity can coexist with predictive performance competitive with a conventional CNN baseline. At the same time, the cross-dataset analysis indicates that computational efficiency alone is not sufficient to explain predictive performance, since ranking stability and final decision quality remained sensitive to dataset characteristics and aggregation strategy.

\section{Theoretical implications for spike-based medical image analysis}

The results of this study provide several insights into the computational role of spiking neural networks in medical imaging. First, the strong performance of the convolutional surrogate-gradient SNN relative to the timing-based Tempotron indicates that spike timing alone is insufficient for complex medical imaging tasks. Instead, explicit convolutional feature extraction appears essential for capturing the spatial structure of abdominal CT data. This finding highlights that spatial feature hierarchies remain a fundamental component even in spike-based models.

Second, the competitive performance of the ConvSNN relative to the CNN baseline suggests that spike-based computation can be effectively applied to high-dimensional visual representations. In particular, the ConvSNN achieved comparable discrimination and strong threshold-dependent performance while operating with sparse internal activity, suggesting that temporal spike-based processing can complement convolutional feature extraction without necessarily sacrificing predictive quality.

More generally, these findings support the view that spiking neural networks should be considered a complementary paradigm to conventional deep learning rather than a direct replacement. Instead of relying solely on dense synchronous activations, SNNs combine sparse event-driven communication with temporal integration of neuronal states. This hybrid representation may be particularly well suited for medical imaging tasks characterized by high spatial resolution, structured anatomical patterns, and limited labeled datasets.

Overall, the results suggest that the most effective spike-based architectures for medical image analysis are those that integrate convolutional spatial representations with temporally dynamic spike-based processing, rather than relying on spike timing alone.

Detailed per-seed performance results are provided in the Supplementary Material (Table~\ref{tab:per_seed_full}).

\section{Conclusion}

This study presented a benchmark of convolutional and spike-based models for CT-based liver lesion presence classification under a strict leakage-free patient-level evaluation protocol across two abdominal CT datasets. A conventional CNN baseline, a convolutional surrogate-gradient SNN, an attention-based ConvSNN-MIL variant, and a Tempotron-style classifier were systematically compared.

The results demonstrate that carefully designed convolutional spiking neural networks can achieve strong and competitive patient-level performance. Although the CNN baseline achieved the highest mean PR-AUC, the surrogate-gradient ConvSNN provided the best overall balance of threshold-dependent metrics, including accuracy, F1-score, balanced accuracy, and MCC. The ConvSNN-MIL variant remained competitive, whereas the timing-based Tempotron model consistently underperformed.

Importantly, the results revealed clear dataset-dependent performance patterns. While convolutional SNNs remained competitive across both datasets, their relative advantage over the CNN baseline varied, highlighting the importance of multi-dataset evaluation in medical imaging benchmarks. This finding suggests that conclusions regarding the methodological value of spike-based models in medical imaging should be based on repeated evaluation across multiple cohorts rather than on isolated single-dataset benchmarks.

These findings indicate that convolutional spike-based feature extraction, strict patient-level evaluation, and validation-based threshold selection are critical for obtaining clinically meaningful results. More broadly, the study supports the view that SNNs can serve as viable and competitive alternatives to conventional deep learning models for CT-based medical image analysis when evaluated under realistic case-level protocols.

Accordingly, the present results should be interpreted as evidence of comparative methodological feasibility under controlled patient-level evaluation rather than as a complete end-to-end clinical deployment study.

Future work should further investigate the relationship between ranking stability, threshold transfer, and calibration under limited patient-level data regimes, as well as validate the proposed approach on larger, multi-center cohorts.

\section*{Conflict of interest/Competing interests}
The authors declare that they have no known competing financial interests or personal relationships that could have appeared to influence the work reported in this paper.

\section*{Data availability}
This study uses two publicly available abdominal CT datasets. The first is the \texttt{Task03\_Liver} dataset from the Medical Segmentation Decathlon (MSD) \cite{antonelli2022}. The second is the multi-phase contrast-enhanced abdominal CT dataset described by Luo et al. \cite{luo2025}, referred to in this study as \texttt{Task03\_CECT}. Both datasets are publicly available for research purposes under their respective access conditions and licenses. Dataset preprocessing and patient-level split generation used in this study are described in the main manuscript.

\section*{Code availability}

A reproducibility-oriented implementation accompanying this study is available at:

\url{https://github.com/PregowskaX/snn-liver-reproducibility}

The repository includes the core model implementations, patient-level evaluation utilities, and a lightweight reviewer-facing toy benchmark that can be executed on CPU without external medical data. The toy benchmark is intended to illustrate the methodological structure of the approach in a fully self-contained setting. The full experimental pipeline depends on the Task03\_Liver dataset from the Medical Segmentation Decathlon and the associated preprocessing steps described in the manuscript.

\section*{Author contribution}
All authors contributed to the conception and design of the study. All authors performed material preparation, data collection, and analysis. The first draft of the manuscript was written by all authors who commented on previous versions of the manuscript. All authors read and approved the final manuscript.

\clearpage
\setcounter{table}{0}
\renewcommand{\thetable}{S\arabic{table}}
\setcounter{figure}{0}
\renewcommand{\thefigure}{S\arabic{figure}}
\setcounter{section}{0}
\renewcommand{\thesection}{S\arabic{section}}

\section*{Supplementary Material}

\subsection*{S1. Per-seed patient-level results}

To ensure full transparency and reproducibility, Tables~\ref{tab:per_seed_full} and~\ref{tab:per_seed_full_cect} report detailed patient-level performance for all evaluated models across the five independent leakage-free patient-level splits.

\begin{table}[htbp]
\centering
\scriptsize
\caption{Per-seed patient-level performance across five leakage-free splits for the \texttt{Task03\_Liver} dataset.}
\label{tab:per_seed_full}
\renewcommand{\arraystretch}{1.15}
\begin{tabular}{lcccccc}
\toprule
Model & Seed & PR-AUC & ROC-AUC & F1 & MCC & Accuracy \\
\midrule

CNN & 0 & 0.946 & 0.947 & 0.800 & 0.617 & 0.808 \\
CNN & 1 & 0.910 & 0.929 & 0.929 & 0.856 & 0.923 \\
CNN & 2 & 1.000 & 1.000 & 0.963 & 0.926 & 0.962 \\
CNN & 3 & 1.000 & 1.000 & 0.929 & 0.856 & 0.923 \\
CNN & 4 & 0.995 & 0.994 & 0.963 & 0.926 & 0.962 \\

\midrule

ConvSNN & 0 & 0.979 & 0.976 & 0.929 & 0.856 & 0.923 \\
ConvSNN & 1 & 0.926 & 0.947 & 0.963 & 0.926 & 0.962 \\
ConvSNN & 2 & 1.000 & 1.000 & 1.000 & 1.000 & 1.000 \\
ConvSNN & 3 & 1.000 & 1.000 & 0.963 & 0.926 & 0.962 \\
ConvSNN & 4 & 0.891 & 0.935 & 0.929 & 0.856 & 0.923 \\

\midrule

ConvSNN-MIL & 0 & 0.900 & 0.941 & 0.960 & 0.889 & 0.960 \\
ConvSNN-MIL & 1 & 0.840 & 0.910 & 0.923 & 0.845 & 0.920 \\
ConvSNN-MIL & 2 & 1.000 & 1.000 & 0.960 & 0.889 & 0.960 \\
ConvSNN-MIL & 3 & 1.000 & 1.000 & 0.960 & 0.889 & 0.960 \\
ConvSNN-MIL & 4 & 0.906 & 0.944 & 0.960 & 0.889 & 0.960 \\

\midrule

Tempotron & 0 & 0.886 & 0.811 & 0.741 & 0.463 & 0.731 \\
Tempotron & 1 & 0.966 & 0.964 & 0.889 & 0.772 & 0.885 \\
Tempotron & 2 & 0.902 & 0.852 & 0.769 & 0.538 & 0.769 \\
Tempotron & 3 & 0.920 & 0.917 & 0.800 & 0.617 & 0.808 \\
Tempotron & 4 & 0.851 & 0.888 & 0.696 & 0.474 & 0.731 \\

\bottomrule
\end{tabular}
\end{table}

\begin{table}[htbp]
\centering
\scriptsize
\caption{Per-seed patient-level performance across five leakage-free splits for the \texttt{Task03\_CECT} dataset.}
\label{tab:per_seed_full_cect}
\renewcommand{\arraystretch}{1.15}
\begin{tabular}{lcccccc}
\toprule
Model & Seed & PR-AUC & ROC-AUC & F1 & MCC & Accuracy \\
\midrule

CNN & 0 & 0.960 & 0.988 & 0.895 & 0.858 & 0.945 \\
CNN & 1 & 0.806 & 0.942 & 0.811 & 0.747 & 0.904 \\
CNN & 2 & 0.862 & 0.936 & 0.722 & 0.633 & 0.863 \\
CNN & 3 & 0.781 & 0.905 & 0.683 & 0.563 & 0.822 \\
CNN & 4 & 0.961 & 0.986 & 0.900 & 0.864 & 0.945 \\

\midrule

ConvSNN & 0 & 0.818 & 0.918 & 0.618 & 0.476 & 0.712 \\
ConvSNN & 1 & 0.716 & 0.840 & 0.744 & 0.648 & 0.849 \\
ConvSNN & 2 & 0.791 & 0.911 & 0.756 & 0.667 & 0.849 \\
ConvSNN & 3 & 0.594 & 0.795 & 0.571 & 0.404 & 0.753 \\
ConvSNN & 4 & 0.910 & 0.936 & 0.850 & 0.795 & 0.918 \\

\midrule

ConvSNN-MIL & 0 & 0.802 & 0.894 & 0.714 & 0.606 & 0.836 \\
ConvSNN-MIL & 1 & 0.659 & 0.823 & 0.684 & 0.573 & 0.836 \\
ConvSNN-MIL & 2 & 0.892 & 0.938 & 0.842 & 0.787 & 0.918 \\
ConvSNN-MIL & 3 & 0.710 & 0.845 & 0.647 & 0.548 & 0.836 \\
ConvSNN-MIL & 4 & 0.959 & 0.949 & 0.944 & 0.929 & 0.973 \\

\midrule

Tempotron & 0 & 0.741 & 0.865 & 0.576 & 0.413 & 0.658 \\
Tempotron & 1 & 0.635 & 0.785 & 0.500 & 0.313 & 0.726 \\
Tempotron & 2 & 0.590 & 0.786 & 0.558 & 0.382 & 0.740 \\
Tempotron & 3 & 0.501 & 0.745 & 0.500 & 0.282 & 0.534 \\
Tempotron & 4 & 0.553 & 0.772 & 0.533 & 0.448 & 0.808 \\

\bottomrule
\end{tabular}
\end{table}

\subsection*{S2. Confidence intervals}

Approximate 95\% confidence intervals (CIs) were estimated across the five independent patient-level splits using a normal approximation. Given the limited number of splits ($n=5$), these intervals should be interpreted as indicative rather than definitive.

\begin{table}[htbp]
\centering
\scriptsize
\caption{Mean performance and approximate 95\% confidence intervals for \texttt{Task03\_Liver} dataset.}
\label{tab:ci}
\renewcommand{\arraystretch}{1.15}
\begin{tabular}{lccc}
\toprule
Model & PR-AUC & F1-score & MCC \\
\midrule
CNN & $0.970$ [0.938, 1.000] & $0.917$ [0.864, 0.970] & $0.836$ [0.724, 0.948] \\
ConvSNN & $0.959$ [0.916, 1.000] & $0.957$ [0.931, 0.983] & $0.913$ [0.860, 0.966] \\
ConvSNN-MIL & $0.929$ [0.865, 0.993] & $0.953$ [0.915, 0.971] & $0.880$ [0.828, 0.932] \\
Tempotron & $0.905$ [0.861, 0.949] & $0.779$ [0.717, 0.841] & $0.573$ [0.477, 0.669] \\
\bottomrule
\end{tabular}
\end{table}

\begin{table}[htbp]
\centering
\scriptsize
\caption{Mean performance and approximate 95\% confidence intervals for the \texttt{Task03\_CECT} dataset.}
\label{tab:ci_cect}
\renewcommand{\arraystretch}{1.15}
\begin{tabular}{lccc}
\toprule
Model & PR-AUC & F1-score & MCC \\
\midrule
CNN & $0.874$ [0.801, 0.948] & $0.802$ [0.716, 0.888] & $0.733$ [0.616, 0.851] \\
ConvSNN & $0.766$ [0.662, 0.870] & $0.708$ [0.610, 0.806] & $0.598$ [0.460, 0.736] \\
ConvSNN-MIL & $0.804$ [0.695, 0.913] & $0.766$ [0.658, 0.875] & $0.689$ [0.545, 0.832] \\
Tempotron & $0.604$ [0.525, 0.684] & $0.534$ [0.504, 0.564] & $0.368$ [0.307, 0.428] \\
\bottomrule
\end{tabular}
\end{table}

These results further emphasize that stability depended on both the metric and the dataset considered. On \texttt{Task03\_Liver}, the standard ConvSNN showed the smallest spread among the high-performing models for ROC-AUC, F1-score, and accuracy, whereas the ConvSNN-MIL model showed particularly low spread for F1-score, MCC, and accuracy but a larger spread for PR-AUC. On \texttt{Task03\_CECT}, the CNN baseline showed relatively low variability for F1-score, MCC, and accuracy compared with the spiking models, although variability remained higher than on the \texttt{Task03\_Liver} dataset.

\begin{table}[htbp]
\centering
\scriptsize
\caption{Observed spread of performance across seeds, computed as max--min for each metric. Lower values indicate greater stability across patient-level splits.}
\label{tab:spread_metrics}
\renewcommand{\arraystretch}{1.15}
\begin{tabular}{lccccc}
\toprule
Model & PR-AUC spread & ROC-AUC spread & F1 spread & MCC spread & Accuracy spread \\
\midrule
CNN & 0.090 & 0.071 & 0.163 & 0.309 & 0.154 \\
ConvSNN & 0.109 & 0.065 & 0.071 & 0.144 & 0.077 \\
ConvSNN-MIL & 0.160 & 0.090 & 0.037 & 0.044 & 0.040 \\
Tempotron & 0.115 & 0.153 & 0.193 & 0.309 & 0.154 \\
\bottomrule
\end{tabular}
\end{table}

\begin{table}[htbp]
\centering
\scriptsize
\caption{Observed spread of performance across seeds for the \texttt{Task03\_CECT} dataset, computed as max--min for each metric. Lower values indicate greater stability across patient-level splits.}
\label{tab:spread_metrics_cect}
\renewcommand{\arraystretch}{1.15}
\begin{tabular}{lccccc}
\toprule
Model & PR-AUC spread & ROC-AUC spread & F1 spread & MCC spread & Accuracy spread \\
\midrule
CNN & 0.179 & 0.083 & 0.217 & 0.301 & 0.123 \\
ConvSNN & 0.316 & 0.141 & 0.279 & 0.391 & 0.206 \\
ConvSNN-MIL & 0.300 & 0.126 & 0.260 & 0.381 & 0.137 \\
Tempotron & 0.240 & 0.120 & 0.076 & 0.166 & 0.274 \\
\bottomrule
\end{tabular}
\end{table}

\subsection*{S3. Threshold analysis}

Validation-selected decision thresholds varied across models and patient-level splits. The CNN baseline exhibited greater threshold variability, suggesting less stable score calibration across partitions. In contrast, the ConvSNN showed more stable threshold selection, consistent with its lower variability in threshold-dependent metrics. This observation further supports the importance of validation-based threshold selection under limited patient-level data regimes. Mean thresholds across seeds are reported in the main manuscript (Table~\ref{tab:thresholds_both}).

\subsection*{S4. Reproducibility}

All per-seed predictions, validation-selected thresholds, and evaluation outputs were saved during experimentation. All experiments were conducted using fixed random seeds and a deterministic patient-level evaluation pipeline. The reported results can be reproduced using the configuration, random seeds, and evaluation protocol described in the main manuscript.

To facilitate reproducibility and provide reviewers with a lightweight executable illustration of the proposed methodology, we include a synthetic CPU-only toy benchmark in the accompanying code repository. This benchmark preserves the key structural elements of the full framework, including bag-level multiple instance learning (MIL), attention-based aggregation, comparison between a CNN baseline and spike-based models, and a simplified time-to-first-spike (TTFS) / Tempotron-style variant.

Unlike the full experimental pipeline, which depends on abdominal CT data, lesion-derived labels, and medical-image preprocessing, the toy benchmark is fully self-contained and requires no external datasets. Its purpose is not to reproduce the medical results reported in this study, but to provide an accessible demonstration of the model logic, training flow, and comparative behavior of the evaluated architectures. Because the synthetic benchmark was intentionally simplified and generated from structured low-dimensional signals, its absolute performance values should not be interpreted as clinically meaningful. In particular, unusually strong performance of the Tempotron-style model in this setting reflects the artificial separability of the synthetic task rather than expected behavior on real medical imaging data.

To improve interpretability, the toy benchmark uses stratified splits, validation-tuned decision thresholds, and multi-seed reporting. Results are summarized in Table~\ref{tab:toy_benchmark}.

\begin{table}[t]
\centering
\scriptsize
\caption{Toy benchmark on synthetic data (CPU-only). Results are reported as mean $\pm$ standard deviation across 3 random seeds using validation-tuned thresholds.}
\begin{tabular}{lccc}
\toprule
Model & ROC-AUC $\uparrow$ & ACC $\uparrow$ & F1 $\uparrow$ \\
\midrule
CNN & $0.993 \pm 0.010$ & $0.933 \pm 0.072$ & $0.942 \pm 0.061$ \\
ConvSNN (SGL) & $0.936 \pm 0.034$ & $0.856 \pm 0.068$ & $0.877 \pm 0.052$ \\
Tempotron (TTFS) & $\mathbf{1.000 \pm 0.000}$ & $\mathbf{0.967 \pm 0.047}$ & $\mathbf{0.970 \pm 0.043}$ \\
\bottomrule
\end{tabular}
\label{tab:toy_benchmark}
\end{table}

We emphasize that this toy benchmark is intended solely as a reproducibility-oriented methodological analogue. It should not be interpreted as a medical validation experiment or as a substitute for the leakage-free patient-level CT evaluation presented in the main study.

\clearpage
\FloatBarrier


\begin{thebibliography}{00}



\bibitem{guo2025}
Z. Guo, D. Wu, R. Mao et al., Global burden of MAFLD, MAFLD related cirrhosis and MASH related liver cancer from 1990 to 2021. Scientific Reports 15 (7083) (2025). https://doi.org/10.1038/s41598-025-91312-5. 

\bibitem{pozowski2025}
P. Pozowski, M. Bilski, M. Bedryło, P. Sitny, U. Zaleska-Dorobisz, Modern ultrasound techniques for diagnosing liver steatosis and fibrosis: A systematic review with a focus on biopsy comparison. World Journal of Hepatology 17(2) (2025) 100033. https://doi.org/10.1038/10.4254/wjh.v17.i2.100033.

\bibitem{hu2023}
N. Hu, G. Yan, M. Tang, Y. Wu, F. Song, X. Xia, L.W. Chan, P. Lei, CT-based methods for assessment of metabolic dysfunction associated with fatty liver disease. European Radiology Experimental 7(1) (2023) 72. https://doi.org/10.1038/10.4254/10.1186/s41747-023-00387-0.
 
\bibitem{chupetlovska2025}
K. Chupetlovska, T. Akinci D’Antonoli, Z. Bodalal, et al., ESR Essentials: a step-by-step guide of segmentation for radiologists - practice recommendations by the European Society of Medical Imaging Informatics. European Radiolog. 35 (2025) 6894--6904. https://doi.org/10.1007/s00330-025-11621-1

\bibitem{ying2024}
H. Ying, X. Liu, M. Zhang M, et al., A multicenter clinical AI system study for detection and diagnosis of focal liver lesions. Nature Communications 15(1) (2024) 1131. https://doi.org/10.1038/s41467-024-45325-9.

\bibitem{he2024}
X. He, Y. Li, D. Zhao, et al., MSAT: biologically inspired multistage adaptive threshold for conversion of spiking neural networks. Neural Comput \& Applic 36 (2024) 8531--8547. https://doi.org/10.1007/s00521-024-09529-w.

\bibitem{xue2023}
X. Xue, R. D. Wimmer,  M. M. Halassa, et al., Spiking Recurrent Neural Networks Represent Task-Relevant Neural Sequences in Rule-Dependent Computation. Cogn Comput 15 (2023) 1167--1189. https://doi.org/10.1007/s12559-022-09994-2.


\bibitem{yang2023}
Y. Yang, J. Ren, F. Duan, The Spiking Rates Inspired Encoder and Decoder for Spiking Neural Networks: An Illustration of Hand Gesture Recognition. Cogn Comput 15 (2023) 1257--1272. https://doi.org/10.1007/s12559-022-10027-1

\bibitem{doborjeh2022}
M. Doborjeh, Z. Doborjeh, A. Merkin A. et al., Personalized Spiking Neural Network Models of Clinical and Environmental Factors to Predict Stroke. Cogn Comput 14 (2022) 2187--2202.  https://doi.org/10.1007/s12559-021-09975-x.


\bibitem{rudnicka2026}
Z. Rudnicka, K. Pauk, J. Pauk, M. Ihnatouski, A. Pregowska, Energy-efficient detection of rheumatoid arthritis using spiking neural networks and thermographic imaging, Biocybernetics and Biomedical Engineering 46 (2026) 266--277. https://doi.org/10.1016/j.bbe.2026.02.004.
    
\bibitem{patankar2025}
M. Patankar, V. Chaurasia, M. Shandilya, A novel spiking neural network method for classification of tuberculosis using X-ray images. Computers and Electrical Engineering 122 (2025) 110003. https://doi.org/10.1016/j.compeleceng.2024.110003.

\bibitem{gilani2023}
S. Q. Gilani, T. Syed, M. Umair, O. Marques, Skin cancer classification using deep spiking neural network. Journal of Digital Imaging 36(3) (2023) 1137--1147. https://doi.org/10.1007/s10278-023-00776-2. 

\bibitem{heidarian2024}
M. Heidarian, G. Karimi, M. Payandeh, Effective multispike learning in a spiking neural network with a new temporal feedback backpropagation for breast cancer detection. Expert Systems with Applications 252(15) (2024) PartB:124010. https://doi.org/10.1016/j.eswa.2024.124010.

\bibitem{dehariya2021}
A. Dehariya, P. Shukla, Medical diagnosis model based on spiking neural network considering normalization of histogram and wavelet transform features of medical images. International Journal of Engineering Trends and Technology 69(7) (2021) 159--166. https://doi.org/10.14445/22315381/IJETT-V69I7P222.

\bibitem{indiveri2009}
G. Indiveri, E. Chicca, R.J. Douglas, Artificial Cognitive Systems: From VLSI Networks of Spiking Neurons to Neuromorphic Cognition. Cogn Comput 1 (2009) 119--127. https://doi.org/10.1007/s12559-008-9003-6.

\bibitem{rudnicka2026neuron}
Z. Rudnicka, J. Szczepanski, A. Pregowska, Impact of Neuron Models on Spiking Neural Network Performance: A Complexity-based Classification Approach, Neuroinformatics 24(5) (2026) 1--23. https://doi.org/10.1007/s12021-025-09754-1.


\bibitem{wu2025}
D. Wu, G. Jin, H. Yu, X. Yi, X. Huan, Optimizing event-driven spiking neural network with regularization and cutoff. Frontiers in Neuroscience 19 (2025) 1522788. https://doi.org/10.3389/fnins.2025.1522788. 

\bibitem{dong2022}
J. F. Dong, R. H. Jiang, R. Xiao, R. Yan, H. J. Tang, Event stream learning using spatio-temporal event surface. Neural Networks 154 (2022) 543--559. https://doi.org/10.1016/j.neunet.2022.07.010.

\bibitem{zhan2023}
Q. Zhan, G. Liu, X Xi.M. Zhan, G. Sun G, Bio-inspired active learning method in spiking neural network. Knowledge-Based Systems 261 (2023) 110193. https://doi.org/10.1016/j.knosys.2022.110193.

\bibitem{zenke2021}
F. Zenke, T.P. Vogels, The remarkable robustness of surrogate gradient learning for instilling complex function in spiking neural networks. Neural Computation 33(4) (2021) 899--925. https://doi.org/10.1162/neco\_a\_01367.

\bibitem{antonelli2022}
M. Antonelli, A. Reinke, S. Bakas, et al., The Medical Segmentation Decathlon. Nature Communications 13 (2022) 4128. https://doi.org/10.1038/s41467-022-30695-9.

\bibitem{luo2025}
J. Luo, X. Wan, J. Du, et al. Comprehensive multi-phase 3D contrast-enhanced CT imaging for primary liver cancer. Sci Data 12, 768 (2025). https://doi.org/10.1038/s41597-025-05125-2.

\bibitem{zhang2025}
X. Dong, Q. Tan, S. Xu, J. Zhang, M. Zhou, Ultrasound image-based contrastive fusion non-invasive liver fibrosis staging algorithm. Abdominal Radiology 50 (2025) 6135--6147 https://doi.org/ 10.1007/s00261-025-04991-z.

\bibitem{kaur2021}
A. Javanshir, T. T. Nguyen, M. A. P. Mahmud, A. Z. Kouzani, Training Spiking Neural Networks with Metaheuristic Algorithms. Appl. Sci. 13 (2023) 4809. https://doi.org/10.3390/app13084809.

\bibitem{lonning2024}
K. Lonning, M. W. A. Caan, M. E. Nowee,  J. J. Sonke, Dynamic recurrent inference machines for accelerated MRI-guided radiotherapy of the liver. Computerized Medical Imaging and Graphics 113 (2024) 102348. https://doi.org/10.1016/j.compmedimag.2024.102348.

\bibitem{yasaka2018}
K. Yasaka, H. Akai, A. Kunimatsu, O. Abe, S. Kiryu, Liver Fibrosis: Deep Convolutional Neural Network for Staging by Using Gadoxetic Acid-enhanced Hepatobiliary Phase MR Images. Radiology 287(1) (2018) 146--155. https://doi.org/ 10.1148/radiol.2017171928.

\bibitem{choi2018}
K.J. Choi, J. K. Jang, S. S. Lee S, et al., Development and Validation of a Deep Learning System for Staging Liver Fibrosis by Using Contrast Agent-enhanced CT Images in the Liver. Radiology 289(3) (2018) 688--697. https://doi.org/10.1148/radiol.2018180763.

\bibitem{yin2021}
Y. Yin, D. Yakar, R.A.J.O. Dierckx, et al., Liver fibrosis staging by deep learning: a visual-based explanation of diagnostic decisions of the model. Eur Radiol 31 (2021) 9620--9627. https://doi.org/10.1007/s00330-021-08046-x.

\end{thebibliography}
\end{document}